\pdfoutput=1
\documentclass[11pt]{article}

\usepackage[preprint]{acl}

\usepackage{times}
\usepackage{latexsym}

\usepackage[T1]{fontenc}
\usepackage[utf8]{inputenc}

\usepackage{microtype}

\usepackage{inconsolata}

\usepackage{graphicx}

\usepackage{amsmath}
\usepackage{multirow}

\usepackage{helvet}
\usepackage{tikz}
\usepackage{pgfplots}
\usepackage{pgfplotstable}
\usepackage{makecell}
\usepackage{booktabs}    
\usetikzlibrary{patterns}

\pgfplotsset{
  compat=1.18,
}
\usepackage{subcaption}  
\usepackage{caption}

\usepackage{xcolor}
\definecolor{myCoral}{HTML}{E99C93}       
\definecolor{myRoseMauve}{HTML}{B88CA9}   
\definecolor{myLavenderBlue}{HTML}{9FACD3} 
\definecolor{myPeach}{HTML}{F1DBB9}        
\definecolor{myLilac}{HTML}{D9D1E3}        
\definecolor{myPeriwinkle}{HTML}{CAD4E7}   
\definecolor{myIceBlue}{HTML}{D0E2E8}      

\definecolor{myHighlightRose}{HTML}{DE6FA1} 
\definecolor{myMintGreen}{HTML}{5DC2B2}
\definecolor{myCrimsonRed}{HTML}{D94F4F}

\definecolor{WarmYellow}{HTML}{FDD0A2}
\definecolor{BrightOrange}{HTML}{FB8E41}
\definecolor{StrongRed}{HTML}{E31B1E}
\definecolor{BrownishOrange}{HTML}{AD5F2C}
\definecolor{BrightYellow}{HTML}{F6E36D}

\definecolor{SoftPink}{HTML}{FAD1E0}
\definecolor{Lavender}{HTML}{D3BAD9}
\definecolor{MediumPurple}{HTML}{B982BC}
\definecolor{MintGreen}{HTML}{CAEAC2}
\definecolor{LightGreen}{HTML}{BBE173}

\definecolor{EmeraldGreen}{HTML}{65C2A4}
\definecolor{SkyBlue}{HTML}{A8CEE0}
\definecolor{MutedBlue}{HTML}{89A6AF}
\definecolor{DeepBlue}{HTML}{1B79AF}

\title{Speculative Decoding Meets Quantization: Compatibility Evaluation and Hierarchical Framework Design}

\author{
 \textbf{Yudi Zhang\textsuperscript{1}},
 \textbf{Weilin Zhao\textsuperscript{2}},
 \textbf{Xu Han\textsuperscript{2}},
 \textbf{Tiejun Zhao\textsuperscript{1}\thanks{\ \ indicates corresponding authors.}},
\\
 \textbf{Wang Xu\textsuperscript{2$^*$}},
 \textbf{Hailong Cao\textsuperscript{1}},
 \textbf{Conghui Zhu\textsuperscript{1}}
\\
 \textsuperscript{1}{Faculty of Computing, Harbin Institute of Technology, Harbin, China.}
 \\
 \textsuperscript{2}{Tsinghua University, Beijing, China.}
\\
 {\tt yudizhang@stu.hit.edu.cn, tjzhao@hit.edu.cn, xwjim812@gmail.com}
}

\begin{document}
\maketitle
\begin{abstract}
Speculative decoding and quantization effectively accelerate memory-bound inference of large language models. Speculative decoding mitigates the memory bandwidth bottleneck by verifying multiple tokens within a single forward pass, which increases computational effort. Quantization achieves this optimization by compressing weights and activations into lower bit-widths and also reduces computations via low-bit matrix multiplications. To further leverage their strengths, we investigate the integration of these two techniques. Surprisingly, experiments applying the advanced speculative decoding method EAGLE-2 to various quantized models reveal that the memory benefits from 4-bit weight quantization are diminished by the computational load from speculative decoding. Specifically, verifying a tree-style draft incurs significantly more time overhead than a single-token forward pass on 4-bit weight quantized models. This finding led to our new speculative decoding design: a hierarchical framework that employs a small model as an intermediate stage to turn tree-style drafts into sequence drafts, leveraging the memory access benefits of the target quantized model. Experimental results show that our hierarchical approach achieves a 2.78$\times$ speedup across various tasks for the 4-bit weight Llama-3-70B model on an A100 GPU, outperforming EAGLE-2 by 1.31$\times$. Code available at \url{https://github.com/AI9Stars/SpecMQuant}.
\end{abstract}

\section{Introduction}

The excellent performance of Large Language Models (LLMs) across diverse domains has driven their widespread integration into everyday applications ~\citep{brown2020language, grattafiori2024llama, guo2025deepseek}. 
However, the large-scale parameters and auto-regressive decoding make inference memory-bound~\citep{patterson2004latency, shazeer2019fast}, particularly on the single-batch inference of resource-constrained devices with model weights dominating memory bandwidth.

To mitigate this memory bandwidth bottleneck, speculative decoding
~\citep{leviathan2023fast, chen2023accelerating} and quantization~\cite{frantar2022gptq, xiao2023smoothquant} are commonly employed in LLM deployment. 
\textbf{\textit{Speculative decoding}} generates multiple tokens in a single forward pass by verifying draft outputs, reducing memory access frequency through increased computation.
Among these methods, self-speculative decoding uses a draft model with the same architecture as the target model~\cite{sun2024triforce, sadhukhan2025magicdec}.
On the other hand, another approach employs a lightweight and independent draft model~\cite{cai2024medusa, li2024eagle1}.
Concurrently, \textbf{\textit{quantization}} improves LLM inference efficiency by reducing memory and computation demands, including weight-only quantization~\citep{frantar2022gptq}, weight-activation quantization~\cite{xiao2023smoothquant}, and KV cache quantization~\cite{hooper2024kvquant}.

Recently, several studies have explored integrating quantization into self-speculative frameworks, whose draft model shares the architecture with the target model but differs in precision.
QSpec~\citep{zhao2024qspec} drafts using 4-bit weights shared with the target model but employing lower-precision 4-bit activations.
QuantSpec~\citep{tiwari2025quantspec} drafts with 4-bit quantized weights and a 4-bit hierarchical KV cache.
ML-SpecQD~\citep{georganas2025ml} similarly employs the 4-bit version of the target model as the draft and further introduces a smaller 4-bit model to enable a multi-level speculative decoding method.

However, self-speculative decoding, which uses the same architecture for both draft and target models, inherently limits speedup.
In contrast, speculative decoding methods with a lightweight draft model achieve superior speedup, as demonstrated by the state-of-the-art approach EAGLE~\cite{li2025eagle}.
To further enhance acceleration and combine the benefits of speculative decoding and quantization, we integrate these two techniques by applying the speculative decoding method with a lightweight draft model to a quantized target model.
Given that speculative decoding and quantization mitigate memory bottlenecks from different perspectives, it is necessary to study their compatibility systematically.
In this paper, we explore two key questions: 
(1) \textbf{How does the integration of speculative decoding and quantization perform?}
Do these two techniques conflict in terms of mitigating memory bottlenecks?
(2) \textbf{Within the integrated framework, what are the dominant factors that affect the overall speedup?}

To achieve the objectives, we first integrated the advanced speculative method EAGLE-2~\citep{li2024eagle} and various optimized quantization kernels into a highly optimized native C and CUDA implementation. 
This implementation filters out non-algorithmic overheads from Python inefficiencies~\citep{zhao2025fr}, thereby revealing each method’s true speedup.
The results of reliable experiments across various integration schemes show that EAGLE-2 provides limited benefit for 4-bit weight quantized models (W4A16 and W4A8), indicating a potential conflict.
Subsequently, we systematically experimented with various draft tree sizes to identify the factors behind the integration conflict.
We find that the increased computational load of tree-style draft verification undermines the memory access benefits from 4-bit weight quantization, leading to limited compatibility.

Motivated by this finding, we propose to design a hierarchical speculative decoding framework for W4A16 quantized models, which have low memory bandwidth demand and near-lossless performance.
To fully leverage the memory advantages of such 4-bit quantized models, we introduce an intermediate stage between EAGLE-2 and the quantized target model, which employs a small model for tree-style verification, turning tree drafts into sequence drafts, enabling fast and accurate drafting without imposing significant verification computational overhead.
Across various tasks with the W4A16 Llama-3-70B model on a single A100 GPU, our hierarchical framework achieves 2.78$\times$ speedup, outperforming the advanced EAGLE-2 method by 1.31$\times$.

\begin{table*}[htbp]
  \centering
  \setlength{\tabcolsep}{4pt}
  \scalebox{0.76}{
  \begin{tabular}{ccllllll}
    \toprule
    \multirow{2}{*}{\textbf{Precision}} & \multirow{2}{*}{\textbf{Algorithm}} & \multicolumn{3}{c}{\textbf{Llama-3-8B-Instruct}} & \multicolumn{3}{c}{\textbf{Llama-3-70B-Instruct}} \\
    \cline{3-8}
      & & WikiText2$\downarrow$  & GSM8K$\uparrow$ & HumanEval$\uparrow$ & WikiText2$\downarrow$ & GSM8K$\uparrow$ & HumanEval$\uparrow$ \\
    \midrule
     FP16 & - & 8.28 & 76.95  & 61.59 & 5.32 & 91.05 & 78.65 \\
    \midrule
     W8A8 & SmoothQuant & 8.37$_{(+0.09)}$ & 77.33$_{(+0.38)}$ & 58.54$_{(-3.05)}$ & 5.87$_{(+0.55)}$ & 90.60$_{(-0.45)}$ & 74.39$_{(-4.26)}$ \\
     \midrule
     \multirow{2}{*}{W4A16} & GPTQ-g128 & 8.73$_{(+0.45)}$  & 73.18$_{(-3.77)}$ & 53.05$_{(-8.54)}$ & 5.86$_{(+0.54)}$ & 89.31$_{(-1.74)}$ & 75.00$_{(-3.65)}$ \\
     & GPTQ-g128+Rot & 8.55$_{(+0.27)}$ & 73.69$_{(-3.26)}$ & 57.93$_{(-3.66)}$ & 5.89$_{(+0.57)}$ & 90.22$_{(-0.83)}$ & 76.22$_{(-2.43)}$ \\
     \midrule
     \multirow{4}{*}{W4A8} & QoQ & 8.73$_{(+0.45)}$ & 73.39$_{(-3.56)}$ & 54.27$_{(-7.32)}$ & 5.97$_{(+0.65)}$ & 88.86$_{(-2.19)}$ & 73.78$_{(-4.87)}$ \\
      & QoQ-g128 & 8.63$_{(+0.35)}$ & 74.07$_{(-2.88)}$ & 56.10$_{(-5.49)}$ & 5.76$_{(+0.44)}$ & 89.69$_{(-1.36)}$ & 73.78$_{(-4.87)}$ \\
      & QQQ & 8.84$_{(+0.56)}$ & 71.65$_{(-5.30)}$ & 50.61$_{(-10.98)}$ & 6.44$_{(+1.12)}$ & 87.57$_{(-3.48)}$ & 73.78$_{(-4.87)}$ \\
      & QQQ-g128 & 8.76$_{(+0.48)}$ & 71.65$_{(-5.30)}$ & 52.44$_{(-9.15)}$ & 6.10$_{(+0.78)}$ & 89.31$_{(-1.74)}$ & 74.39$_{(-4.26)}$ \\
    \bottomrule
  \end{tabular}
  }  
  \caption{WikiText2 perplexity with 2048 sequence length, 8-shot performance on GSM8K and zero-shot performance on HumanEval of different quantized method on Llama-3-8B-Instruct and Llama-3-70B-Instruct.}
  \label{tab:quant_performance}
  \vspace{-5pt}
\end{table*}

\section{Preliminary}
In this section, we present the speculative decoding method EAGLE-2, the quantization schemes (W8A8, W4A16, and W4A8) investigated in this paper, and the performance of quantized models.

\subsection{Speculative Decoding}
Speculative Decoding presents a draft-then-verify decoding paradigm. 
EAGLE-2 uses a lightweight module that consists of a single Transformer layer and uses the original LM Head to generate tree-style draft tokens auto-regressively. 
It dynamically adjusts the draft tree structure by adopting beam-search algorithm based on the softmax output of the draft model.
During drafting, the draft model forwards $d$ times and selects the top $n$ probability tokens from the beam search history as the draft, where $d$ is the search depth and $n$ is the tree size. 

Let $\tau(n, d)$ denote the expected accepted length, defined as the expected number of tokens accepted by the target model after verifying the drafts.
$T_d$ and $T_t$ denote the decoding time of the draft model and target model, respectively. $T_v(n)$ denotes the time taken by the target model to verify $n$ tokens. $T^{sd}_{avg}$ represents the expected latency per token for speculative decoding.
The speedup effect of speculative decoding can be understood with the following equation~\citep{sadhukhan2025magicdec}:
\begin{equation}
  \frac{T^{sd}_{avg}}{T_{t}} = \frac{1}{\tau(n, d)} \left( \frac{d \cdot T_d}{T_t} + \frac{T_v(n)}{T_t} \right)
  \label{eq:speedup}
\end{equation}

The impressive speedup achieved by EAGLE-2 is mainly attributed to three factors:(1) a high excepted accepted length $\tau(n, d)$, (2) a low draft-to-target decoding time ratio $T_d / T_t$ close to 0, (3) a low target verification-to-decoding time ratio $T_{v}(n)/T_t$ close to 1.
In addition, these three factors also guide our subsequent speed analysis when combined with quantization.

\subsection{Quantization}

Quantization methods compress model weights and activations into low-bit representations.
In this paper, we denote x-bit weight and y-bit activation quantization precision in LLM as WxAy.
The following is a brief introduction of quantization precisions and algorithms investigated in this paper:

\textit{\textbf{W8A8:}} For 8-bit weight-activation quantization, SmoothQuant~\citep{xiao2023smoothquant} shifts the quantization difficulty from activations to weights. We adopted SmoothQuant with channel-wise scaling in our W8A8 experiments. This method leverages INT8 Tensor Cores to lower computation costs.
  
\textit{\textbf{W4A16:}} GPTQ~\citep{frantar2022gptq} adopts second-order information to minimize precision loss for weight-only quantization. We adopted GPTQ to symmetrically quantize weights to 4-bit with a group size of 128 while keeping activation 16-bit. This method mitigates memory bottleneck.
  
\textit{\textbf{W4A8:}}To achieve 4-bit weights and 8-bit activations, QoQ~\citep{lin2024qserve} employs an asymmetric scheme with better accuracy performance and QQQ~\citep{zhang2024qqq} employs a symmetric quantization scheme for superior efficiency.
They support 4-bit weight quantization with both per-channel and per-group granularity, while enabling matrix multiplications on INT8 Tensor Cores.

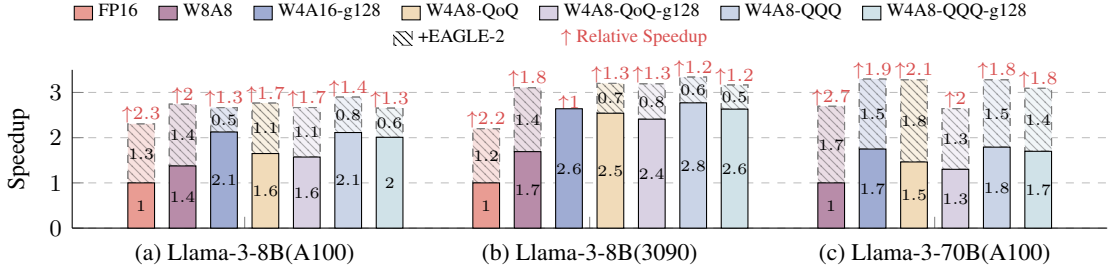
\begin{figure*}[htbp]
  \centering
  \begin{tikzpicture}
    \begin{axis}[
        hide axis,
        width=0.95\linewidth,
        height=0.1\linewidth,
        xmin=0, xmax=1, ymin=0, ymax=1,
        legend style={
          font=\scriptsize,
          legend columns=7,
          /tikz/every even column/.append style={column sep=0.2cm},
          draw=none,
          fill=none,
          inner sep=0pt,
          at={(0.0,1.0)}, anchor=center
        },
        legend image code/.code={
          \draw[#1,fill=#1] (0cm,-0.1cm) rectangle (0.24cm,0.08cm);
        }
    ]

    \addlegendimage{fill=myCoral}
    \addlegendentry{FP16}

    \addlegendimage{fill=myRoseMauve}
    \addlegendentry{W8A8}

    \addlegendimage{fill=myLavenderBlue}
    \addlegendentry{W4A16-g128}

    \addlegendimage{fill=myPeach}
    \addlegendentry{W4A8-QoQ}
    
    \addlegendimage{fill=myLilac}
    \addlegendentry{W4A8-QoQ-g128}

    \addlegendimage{fill=myPeriwinkle} 
    \addlegendentry{W4A8-QQQ}

    \addlegendimage{fill=myIceBlue}
    \addlegendentry{W4A8-QQQ-g128}

    \end{axis}
    \begin{axis}[
        at={(0,-0.02\linewidth)}, anchor=north west,
        hide axis, 
        width=0.95\linewidth,
        height=0.1\linewidth,
        xmin=0, xmax=1, ymin=0, ymax=1,
        legend style={
          font=\scriptsize,
          legend columns=7,
          /tikz/every even column/.append style={column sep=0.6cm},
          draw=none,
          fill=none,
          inner sep=0pt,
          at={(-0.00,1.0)}, anchor=center
        },
        legend image code/.code={
          \draw[#1,fill=#1] (0cm,-0.1cm) rectangle (0.24cm,0.08cm);
        }
    ]

    \addlegendimage{
      fill=white,
      fill opacity=1,
      draw=gray,
      dashed,
      postaction={pattern=north west lines}
    }
    \addlegendentry{  +EAGLE-2}

    \addlegendimage{empty legend}  

    \addlegendentry{{\scriptsize \textcolor{myCrimsonRed}{$\uparrow$ Relative Speedup}}} 

    \end{axis}
  \end{tikzpicture}

    \begin{tikzpicture}
    \begin{axis}[
        ybar stacked,
        name=main,
        width=0.95\linewidth,
        height=0.23\linewidth,
        axis lines*=left,
        ymin=0, ymax=3.5,
        ylabel={Speedup},
        xmin=-0.5, xmax=2.5,
        xtick={0,1,2},
        xticklabels={(a) Llama-3-8B(A100), (b) Llama-3-8B(3090), (c) Llama-3-70B(A100)},
        xticklabel style={font=\small, yshift=-2pt},
        ylabel style={font=\small},
        yticklabel style={font=\small},
        ymajorgrids=true,
        grid style=dashed,
        bar width=10pt,
        nodes near coords,
        nodes near coords style={
          font=\tiny,
          inner sep=1pt,
          /pgf/number format/.cd,
          fixed,
          precision=1,
        },
        legend style={ 
          font=\scriptsize,  
          at={(0.5,1.05)},
          anchor=south,
          legend columns=7,
          /tikz/every even column/.append style={column sep=0.2cm},
          draw=none,         
          fill=none,         
          inner sep=0pt,     
        },
        legend image code/.code={
          \draw[#1,fill=#1] (0cm,-0.1cm) rectangle (0.24cm,0.08cm);
        },
    ]

    \addplot[ybar, fill=myCoral] plot coordinates { (-0.30,1) (0.70,1)}; 

    \addplot[ybar,fill=myCoral!30, draw=gray, dashed, postaction={pattern=north west lines, fill opacity=0.3}] plot coordinates {(-0.30, 1.3054998665481254) (0.70, 1.199199431692463)};

    \node[
      font=\scriptsize\bfseries\itshape,
      text=myCrimsonRed,
      rounded corners=2pt,
      inner sep=2pt
    ] at (axis cs:-0.30, 2.3054998665481254+0.2) {$\uparrow$\pgfmathprintnumber[fixed, precision=1]{2.3054998665481254}};
    \node[
      font=\scriptsize\bfseries\itshape,
      text=myCrimsonRed,
      rounded corners=2pt,
      inner sep=2pt
    ] at (axis cs:0.70, 1+1.199199431692463+0.2) {$\uparrow$\pgfmathprintnumber[fixed, precision=1]{2.199199431692463}};
    \end{axis}

    \begin{axis}[
      ybar stacked,
      name=main,
      width=0.95\linewidth,
      height=0.23\linewidth,
      axis y line=none,
      axis x line=none,
      xmin=-0.5, xmax=2.5,
      ymin=0, ymax=3.5,
      ymajorgrids=true,
      bar width=10pt,
      nodes near coords,
      nodes near coords style={
        font=\tiny,
        inner sep=1pt,
        /pgf/number format/.cd,
        fixed,
        precision=1
      },
      legend style={ 
        font=\scriptsize,  
        at={(0.5,1.05)},
        anchor=south,
        legend columns=7,
        /tikz/every even column/.append style={column sep=0.2cm},
        draw=none,         
        fill=none,         
        inner sep=0pt,     
      },
      legend image code/.code={
        \draw[#1,fill=#1] (0cm,-0.1cm) rectangle (0.24cm,0.08cm);
      },
  ]

  \addplot[ybar, fill=myRoseMauve] coordinates {(-0.18, 1.3747246327414215) (0.82, 1.6912597160360925) (1.70, 1.0)};

  \addplot[ybar, fill=myRoseMauve!30, draw=gray, dashed, postaction={pattern=north west lines, fill opacity=0.3}] coordinates {(-0.18, 2.7425303258567064-1.3747246327414215) (0.82, 3.1051827130209264- 1.6912597160360925) (1.70, 2.695243434661454 - 1.0)}; 

  \node[
      font=\scriptsize\bfseries\itshape,
      text=myCrimsonRed,
      rounded corners=2pt,
      inner sep=2pt
    ] at (axis cs:-0.18, 2.7425303258567064+0.2) {$\uparrow$\pgfmathprintnumber[fixed, precision=1]{1.994967035971168}};
    \node[
      font=\scriptsize\bfseries\itshape,
      text=myCrimsonRed,
      rounded corners=2pt,
      inner sep=2pt
    ] at (axis cs:0.82, 3.1051827130209264+0.2) {$\uparrow$\pgfmathprintnumber[fixed, precision=1]{1.8360176639805097}};
    \node[
      font=\scriptsize\bfseries\itshape,
      text=myCrimsonRed,
      rounded corners=2pt,
      inner sep=2pt
    ] at (axis cs:1.70, 2.695243434661454+0.2) {$\uparrow$\pgfmathprintnumber[fixed, precision=1]{2.695243434661454}};

  \end{axis}

  \begin{axis}[
    ybar stacked,
    name=main,
    width=0.95\linewidth,
    height=0.23\linewidth,
    axis y line=none,
    axis x line=none,
    xmin=-0.5, xmax=2.5,
    ymin=0, ymax=3.5,
    ymajorgrids=true,
    bar width=10pt,
    clip=false,  
    nodes near coords,
    nodes near coords style={
      font=\tiny,
      inner sep=1pt,
      /pgf/number format/.cd,
      fixed,
      precision=1
    },
    legend style={ 
      font=\scriptsize,  
      at={(0.5,1.05)},
      legend columns=7,
      /tikz/every even column/.append style={column sep=0.2cm},
      draw=none,         
      fill=none,         
      inner sep=0pt,     
    },
    legend image code/.code={
      \draw[#1,fill=#1] (0cm,-0.1cm) rectangle (0.24cm,0.08cm);
    },
]

\addplot[ybar, fill=myLavenderBlue] coordinates {(-0.06, 2.1250626900091043) (0.94, 2.6395548157781574) (1.82,1.7476507090842306)}; 
\addplot[ybar, fill=myLavenderBlue!30, draw=gray, dashed, postaction={pattern=north west lines, fill opacity=0.3}] coordinates {(-0.06, 2.6648778328104377-2.1250626900091043) (0.94, 0) (1.82, 3.297596779230136-1.7476507090842306)}; % W4A16

  \node[
    font=\scriptsize\bfseries\itshape,
    text=myCrimsonRed,
    rounded corners=2pt,
    inner sep=2pt
  ] at (axis cs:-0.06, 2.6648778328104377+0.2) {$\uparrow$\pgfmathprintnumber[fixed, precision=1]{1.2540231614527195}};
  \node[
    font=\scriptsize\bfseries\itshape,
    text=myCrimsonRed,
    rounded corners=2pt,
    inner sep=2pt
  ] at (axis cs:0.94, 2.569033514965405+0.2) {$\uparrow$\pgfmathprintnumber[fixed, precision=1]{0.9732828807376132}};
  \node[
    font=\scriptsize\bfseries\itshape,
    text=myCrimsonRed,
    rounded corners=2pt,
    inner sep=2pt
  ] at (axis cs:1.82, 3.297596779230136+0.2) {$\uparrow$\pgfmathprintnumber[fixed, precision=1]{1.8868740544602747}};

\end{axis}

\begin{axis}[
  ybar stacked,
  name=main,
  width=0.95\linewidth,
  height=0.23\linewidth,
  axis y line=none,
  axis x line=none,
  xmin=-0.5, xmax=2.5,
  ymin=0, ymax=3.5,
  ymajorgrids=true,
  bar width=10pt,
  clip=false, 
  nodes near coords,
  nodes near coords style={
    font=\tiny,
    inner sep=1pt,
    /pgf/number format/.cd,
    fixed,
    precision=1
  },
  legend style={ 
    font=\scriptsize,  
    at={(0.5,1.05)},
    anchor=south,
    legend columns=7,
    /tikz/every even column/.append style={column sep=0.2cm},
    draw=none,         
    fill=none,         
    inner sep=0pt,     
  },
  legend image code/.code={
    \draw[#1,fill=#1] (0cm,-0.1cm) rectangle (0.24cm,0.08cm);
  },
]

\addplot[bar shift=0, fill=myPeach] coordinates {(0.06, 1.6488987796832808) (1.06,2.540211317062392) (1.94,1.4624872706840533)};

\addplot[bar shift=0, fill=myPeach!40, draw=gray, dashed, postaction={pattern=north west lines, fill opacity=0.4}] coordinates {(0.06, 2.766534092998038-1.6488987796832808) (1.06, 3.200545022486774-2.540211317062392) (1.94,3.2811463167988655-1.4624872706840533)}; 

\node[
  font=\scriptsize\bfseries\itshape,
  text=myCrimsonRed,
  rounded corners=2pt,
  inner sep=2pt
] at (axis cs:0.06, 2.766534092998038+0.2) {$\uparrow$\pgfmathprintnumber[fixed, precision=1]{1.6778071080442134}};
\node[
  font=\scriptsize\bfseries\itshape,
  text=myCrimsonRed,
  rounded corners=2pt,
  inner sep=2pt
] at (axis cs:1.06, 3.200545022486774+0.2) {$\uparrow$\pgfmathprintnumber[fixed, precision=1]{1.2599522728636687}};
\node[
  font=\scriptsize\bfseries\itshape,
  text=myCrimsonRed,
  rounded corners=2pt,
  inner sep=2pt
] at (axis cs:1.94, 3.2811463167988655+0.2) {$\uparrow$\pgfmathprintnumber[fixed, precision=1]{2.0848542148415397}};

\end{axis}

\begin{axis}[
  ybar stacked,
  name=main,
  width=0.95\linewidth,
  height=0.23\linewidth,
  axis y line=none,
  axis x line=none,
  xmin=-0.5, xmax=2.5,
  ymin=0, ymax=3.5,
  ymajorgrids=true,
  bar width=10pt,
  clip=false,  
  nodes near coords,
  nodes near coords style={
    font=\tiny,
    inner sep=1pt,
    /pgf/number format/.cd,
    fixed,
    precision=1
  },
  legend style={ 
    font=\scriptsize,  
    at={(0.5,1.05)},
    anchor=south,
    legend columns=7,
    /tikz/every even column/.append style={column sep=0.2cm},
    draw=none,         
    fill=none,         
    inner sep=0pt,     
  },
  legend image code/.code={
    \draw[#1,fill=#1] (0cm,-0.1cm) rectangle (0.24cm,0.08cm);
  },
]

\addplot[ybar, fill=myLilac] coordinates {(0.18, 1.57137768419888) (1.18, 2.4082815138277516) (2.06, 1.2999586910317988)}; 

\addplot[ybar, fill=myLilac!30, draw=gray, dashed, postaction={pattern=north west lines, fill opacity=0.3}] coordinates {(0.18, 2.666772226427491-1.57137768419888) (1.18, 3.1924234043704276-2.4082815138277516) (2.06, 2.6430648868195603-1.2999586910317988)}; 

\node[
  font=\scriptsize\bfseries\itshape,
  text=myCrimsonRed,
  rounded corners=2pt,
  inner sep=2pt
] at (axis cs:0.18, 2.666772226427491+0.2) {$\uparrow$\pgfmathprintnumber[fixed, precision=1]{1.697091827918547}};
\node[
  font=\scriptsize\bfseries\itshape,
  text=myCrimsonRed,
  rounded corners=2pt,
  inner sep=2pt
] at (axis cs:1.18, 3.1924234043704276+0.2) {$\uparrow$\pgfmathprintnumber[fixed, precision=1]{1.3256022545704598}};
\node[
  font=\scriptsize\bfseries\itshape,
  text=myCrimsonRed,
  rounded corners=2pt,
  inner sep=2pt
] at (axis cs:2.06, 2.6430648868195603+0.2) {$\uparrow$\pgfmathprintnumber[fixed, precision=1]{2.033191442969404}};

\end{axis}

\begin{axis}[
  ybar stacked,
  name=main,
  width=0.95\linewidth,
  height=0.23\linewidth,
  axis y line=none,
  axis x line=none,
  xmin=-0.5, xmax=2.5,
  ymin=0, ymax=3.5,
  ymajorgrids=true,
  bar width=10pt,
  clip=false,  
  nodes near coords,
  nodes near coords style={
    font=\tiny,
    inner sep=1pt,
    /pgf/number format/.cd,
    fixed,
    precision=1
  },
  legend style={ 
    font=\scriptsize,  
    at={(0.5,1.05)},
    anchor=south,
    legend columns=7,
    /tikz/every even column/.append style={column sep=0.2cm},
    draw=none,         
    fill=none,         
    inner sep=0pt,     
  },
  legend image code/.code={
    \draw[#1,fill=#1] (0cm,-0.1cm) rectangle (0.24cm,0.08cm);
  },
]

\addplot[bar shift=0, fill=myPeriwinkle] coordinates {(0.30,2.1136596398217584) (1.30, 2.7699827198506193) (2.18,1.7901435001767059)};

\addplot[bar shift=0, fill=myPeriwinkle!30, draw=gray, dashed, postaction={pattern=north west lines, fill opacity=0.3}] coordinates {(0.30, 2.8991776813384402-2.1136596398217584) (1.30, 3.3415109640287426-2.7699827198506193) (2.18,3.2811463167988655-1.7901435001767059)}; 

\node[
  font=\scriptsize\bfseries\itshape,
  text=myCrimsonRed,
  rounded corners=2pt,
  inner sep=2pt
] at (axis cs:0.30, 2.8991776813384402+0.2) {$\uparrow$\pgfmathprintnumber[fixed, precision=1]{1.3716388517419595}};
\node[
  font=\scriptsize\bfseries\itshape,
  text=myCrimsonRed,
  rounded corners=2pt,
  inner sep=2pt
] at (axis cs:1.30, 3.3415109640287426+0.2) {$\uparrow$\pgfmathprintnumber[fixed, precision=1]{1.2063291731325112}};
\node[
  font=\scriptsize\bfseries\itshape,
  text=myCrimsonRed,
  rounded corners=2pt,
  inner sep=2pt
] at (axis cs:2.18, 3.2811463167988655+0.2) {$\uparrow$\pgfmathprintnumber[fixed, precision=1]{1.832895696057317}};

\end{axis}

\begin{axis}[
  ybar stacked,
  name=main,
  width=0.95\linewidth,
  height=0.23\linewidth,
  axis y line=none,
  axis x line=none,
  xmin=-0.5, xmax=2.5,
  ymin=0, ymax=3.5,
  ymajorgrids=true,
  bar width=10pt,
  clip=false,  
  nodes near coords,
  nodes near coords style={
    font=\tiny,
    inner sep=1pt,
    /pgf/number format/.cd,
    fixed,
    precision=1
  },
  legend style={ 
    font=\scriptsize,  
    at={(0.5,1.05)},
    anchor=south,
    legend columns=7,
    /tikz/every even column/.append style={column sep=0.2cm},
    draw=none,         
    fill=none,         
    inner sep=0pt,     
  },
  legend image code/.code={
    \draw[#1,fill=#1] (0cm,-0.1cm) rectangle (0.24cm,0.08cm);
  },
]

\addplot[bar shift=0, fill=myIceBlue] coordinates {(0.42, 2.0077669742145656) (1.42, 2.6304927884500584) (2.30, 1.6983789893150953)}; 

\addplot[bar shift=0, fill=myIceBlue!30, draw=gray, dashed, postaction={pattern=north west lines, fill opacity=0.3}] coordinates {(0.42, 2.654831594025623-2.0077669742145656) (1.42, 3.1667240181690137-2.6304927884500584) (2.30, 3.0923810722867144-1.6983789893150953)};

\node[
  font=\scriptsize\bfseries\itshape,
  text=myCrimsonRed,
  rounded corners=2pt,
  inner sep=2pt
] at (axis cs:0.42, 2.654831594025623+0.2) {$\uparrow$\pgfmathprintnumber[fixed, precision=1]{1.322280736819166}};
\node[
  font=\scriptsize\bfseries\itshape,
  text=myCrimsonRed,
  rounded corners=2pt,
  inner sep=2pt
] at (axis cs:1.42, 3.1667240181690137+0.2) {$\uparrow$\pgfmathprintnumber[fixed, precision=1]{1.2038520052491435}};
\node[
  font=\scriptsize\bfseries\itshape,
  text=myCrimsonRed,
  rounded corners=2pt,
  inner sep=2pt
] at (axis cs:2.30, 3.0923810722867144+0.2) {$\uparrow$\pgfmathprintnumber[fixed, precision=1]{1.820783871998898}};

\end{axis}
 
    \end{tikzpicture}
    \caption{Comparison of speedup ratios for Llama-3-8B (relative to FP16) and Llama-3-70B (relative to W8A8) under various quantization methods and with EAGLE-2 integration. Solid bars show speedup from quantization alone, dashed bars represent the additional speedup from EAGLE-2, and red arrows indicate the relative speedup achieved by EAGLE-2 across different quantized models.}
    \label{fig:diff_quant_ea_speedup}
   \vspace{-5pt}
  \end{figure*}

In both W4A8 quantization methods, rotation with Hadamard transformation~\citep{ashkboos2024quarot} was introduced as a general offline quantization optimization. 
Inspired by ~\citet{sun2024flatquant} demonstrating that Hadamard transformation can improve the flatness of model weights, we also applied rotation optimization to the W4A16 quantization.
Additionally, a subset of 128 sequences sampled from the Pile validation dataset~\cite{gao2020pile} was used for calibration.

To evaluate the performance of different quantization precisions, we conduct experiments on Llama-3-8B-Instruct and Llama-3-70B-Instruct models~\cite{grattafiori2024llama} quantized with various algorithms.
Three benchmarks are evaluated: WikiText2~\citep{merity2016pointer} for perplexity, GSM8K~\citep{cobbe2021training} for arithmetic reasoning, and HumanEval~\citep{chen2021evaluating} for code generation.
As shown in Table~\ref{tab:quant_performance}, W8A8 and W4A16 achieve the best performance and maintain near-lossless performance, followed by the asymmetric W4A8 quantization method (QoQ), while the symmetric W4A8 algorithm (QQQ) exhibits the most significant degradation.

\section{Experimental Study for Integration}

In this section, we first present the setup and our experimental results in Section~\ref{subsec:experimet_setup} and Section~\ref{subsec:comparison_speedup}, respectively.
While integration improves overall speedup, applying EAGLE-2 on 4-bit weight quantized models (W4A16 and W4A8) yields limited additional speedups compared to higher precision settings.
To investigate this limitation, Section~\ref{subsec:eagle_analysis} explores the underlying factors that hinder further memory bandwidth optimization through experiments with varying draft tree sizes.

\subsection{Experimental Setup}
\label{subsec:experimet_setup}

\textbf{Models and Datasets.} 
We conduct experiments on Llama-3 series models~\citep{grattafiori2024llama}, including Llama-3-8B-Insturct and Llama-3-70B-Instruct. And we evaluate the decoding speedup of different methods using the multi-turn conversation dataset MT-Bench~\citep{zheng2023judging}.

\textbf{Speculative Decoding.} 
We adopt EAGLE-2 ~\citep{li2024eagle} as the speculative method and the implementation in native C and CUDA ~\citep{zhao2025fr}.
Following the original settings of EAGLE-2, we set the search depth $d$ to 6 and the tree size $n$ to 60 for the Llama-3-8B-Instruct model, and a search depth $d$ of 6 and a tree size $n$ of 48 for the Llama-3-70B-Instruct model. 
And we keep the draft model FP16 precision due to the main bottleneck of drafting is the language model head and softmax operation, which are not quantizable ~\cite{zhao2025fr}. Moreover, GPTQ quantization to the draft model leads to substantial degradation of the acceptance rate~\citep{zhao2024qspec}.

\textbf{Quantization Methods.} 
We evaluate several representative quantization methods: W8A8, W4A16, W4A8-QoQ, W4A8-QoQ-g128, W4A8-QQQ, and W4A8-QQQ-g128. 
For W4A16, we use advanced Marlin kernels~\citep{frantar2025marlin} adapted by vLLM~\citep{kwon2023efficient}. 
For W8A8, W4A8-QoQ, and W4A8-QoQ-g128, we adopt the implementation from QServe~\citep{lin2024qserve}. 
For W4A8-QQQ and W4A8-QQQ-g128, we employ novel kernels from QQQ~\citep{zhang2024qqq}.

\textbf{Hardware.} 
Experiments are conducted on a single NVIDIA 80G A100 and a single RTX 3090, representing high-performance and consumer-grade GPUs, respectively.

\begin{figure*}[t]
  \centering
  \begin{tikzpicture}
    \begin{axis}[
      hide axis,
      xmin=0,xmax=1,ymin=0,ymax=1,
      legend style={
        draw=black,      
        font=\scriptsize,
        legend columns=-1,
        at={(0.5,1)},   
        anchor=south,
        draw=none,        
        fill=none,        
      }
    ]
    \addlegendimage{StrongRed,mark=*}
    \addlegendimage{MediumPurple,mark=*}
    \addlegendimage{DeepBlue,mark=*}
    \addlegendimage{EmeraldGreen,mark=*}
    \addlegendimage{LightGreen,mark=*}
    \legend{FP16, W8A8, W4A16, W4A8-QQQ, W4A8-QQQ-g128}
    \end{axis}
  \end{tikzpicture}

  \begin{subfigure}[b]{0.32\textwidth}
    \centering
    \begin{tikzpicture}
        \begin{axis}[
            xlabel={\scriptsize Draft tree size},
            ylabel={\scriptsize Avg. accepted length},
            xmin=28, xmax=62,
            ymin=3, ymax=4.3,
            xtick={30,40,50,60},
            grid=major,
            width=\textwidth,
            height=0.75\textwidth,
            tick label style={font=\tiny},
            legend style={font=\tiny, at={(0.5,1.05)}, anchor=south, legend columns=-1},
        ]
        \addplot[color=StrongRed,mark=*] coordinates {(30, 3.356260904576567)(40, 3.727159316063697)(50, 3.9882510645008673)(60, 4.118646837718582)};
        \addplot[color=MediumPurple,mark=*] coordinates {(30, 3.3348413322842907)(40, 3.699142291620849)(50, 3.9279899812147776)(60, 4.05714748259673)}; 
        \addplot[color=DeepBlue,mark=*] coordinates {(30, 3.299505440158259)(40, 3.6693341164378532)(50, 3.9022474531456566)(60, 4.046442129068734)};
        \addplot[color=EmeraldGreen,mark=*] coordinates {(30, 3.2729444555046783)(40, 3.611221853082318)(50, 3.8006528142663)(60, 3.928245375190198)}; 
        \addplot[color=LightGreen, mark=*] coordinates {(30, 3.2562113759616595)(40,3.563006020344613)(50, 3.7967827626918536)(60,3.9141575425790753)}; 
        \end{axis}
    \end{tikzpicture}
    \caption{Accepted Length (8B, A100)}
    \label{fig:8b_a100_accept}
\end{subfigure}
\hfill
\begin{subfigure}[b]{0.32\textwidth}
    \centering
    \begin{tikzpicture}
        \begin{axis}[
            xlabel={\scriptsize Draft tree size},
            ylabel={\scriptsize Verification/decoding},
            xmin=28, xmax=62,
            ymin=1, ymax=2,
            xtick={30,40,50,60},
            grid=major,
            width=\textwidth,
            height=0.75\textwidth,
            tick label style={font=\tiny},
            legend style={font=\tiny},
            legend pos=south west
        ]
        \addplot[color=StrongRed,mark=*] coordinates {(30, 1.1316109749639263)(40, 1.1185495929520781)(50, 1.063201195522366)(60, 1.143818338918584)}; 
        \addplot[color=MediumPurple,mark=*] coordinates {(30, 1.091813309181821)(40, 1.1066786677659897)(50, 1.126892027901156)(60, 1.1437940038364323)}; 
        \addplot[color=DeepBlue,mark=*] coordinates {(30, 1.3172193621103474)(40, 1.5475202899541272)(50, 1.7900923637762565)(60, 1.8033141240416124)};
        \addplot[color=EmeraldGreen,mark=*] coordinates {(30, 1.2461308184365094)(40, 1.3701383891966064)(50, 1.4454781090009785)(60, 1.472112194507588)}; 
        \addplot[color=LightGreen, mark=*] coordinates {(30, 1.2698140969254474)(40, 1.454571446934599)(50, 1.6175741033302469)(60, 1.6673240510080822)};  
        \end{axis}
    \end{tikzpicture}
    \caption{Verification Ratio (8B, A100)}
    \label{fig:8b_a100_verification_ratio}
\end{subfigure}
\hfill
  \begin{subfigure}[b]{0.32\textwidth}
    \centering
    \begin{tikzpicture}
        \begin{axis}[
            xlabel={\scriptsize Draft tree size},
            ylabel={\scriptsize Speedup},
            xmin=28, xmax=62,
            ymin=2, ymax=3.5,
            xtick={30,40,50,60},
            ytick={2, 2.5, 3, 3.5},
            grid=major,
            width=\textwidth,
            height=0.75\textwidth,
            tick label style={font=\tiny},
        ]
        \addplot[color=StrongRed,mark=*] coordinates {(30, 2.281886166758471)(40, 2.3832322733783884)(50, 2.4639673099324937)(60, 2.3054998665481254)}; 
        \addplot[color=MediumPurple,mark=*] coordinates {(30, 2.9282758777691202)(40, 2.9469532134450853)(50, 2.871356239174034)(60, 2.7425303258567064)}; 
        \addplot[color=DeepBlue,mark=*] coordinates {(30, 3.396056945230414)(40, 3.081891759207875)(50, 2.7761315336373613)(60, 2.6648778328104377)}; 
        \addplot[color=EmeraldGreen,mark=*] coordinates {(30, 3.470039366278271)(40, 3.2562810865692975)(50, 3.054162443915391)(60, 2.8991776813384402)}; 
        \addplot[color=LightGreen, mark=*] coordinates {(30, 3.3498371442427017)(40, 3.075373986743785)(50, 2.832752403680478)(60,2.654831594025623)};  
        \end{axis}
    \end{tikzpicture}
    \caption{Speedup (8B, A100)}
    \label{fig:8b_a100_speedup}
\end{subfigure}

\begin{subfigure}[b]{0.32\textwidth}
  \centering
  \begin{tikzpicture}
      \begin{axis}[
          xlabel={\scriptsize Draft tree size},
          ylabel={\scriptsize Avg. accepted length},
          xmin=28, xmax=62,
          ymin=3, ymax=4.3,
          xtick={30,40,50,60},
          grid=major,
          width=\textwidth,
          height=0.75\textwidth,
          tick label style={font=\tiny},
          legend style={font=\tiny},
          legend pos=south west
      ]
      \addplot[color=StrongRed,mark=*] coordinates {(30, 3.34786059123073)(40, 3.726936723832052)(50, 3.9726178298843338)(60, 4.128497579387872)}; 
      \addplot[color=MediumPurple,mark=*] coordinates {(30, 3.3294563512807107)(40, 3.686229749631811)(50, 3.9078740157480314)(60, 4.04457301173402)}; 
      \addplot[color=DeepBlue,mark=*] coordinates {(30, 3.2951451152287974)(40, 3.6580188679245285)(50, 3.9098226146753143)(60, 4.043467643467643)}; 
      \addplot[color=EmeraldGreen,mark=*] coordinates {(30, 3.263549555482384)(40, 3.620608899297424)(50, 3.7921472586506555)(60, 3.9434261683726097)}; 
      \addplot[color=LightGreen, mark=*] coordinates {(30, 3.262012578616352)(40, 3.5663967891495396)(50, 3.7821687807023974)(60, 3.8935234339450933)};  
      \end{axis}
  \end{tikzpicture}
  \caption{Accepted Length (8B, 3090)}
  \label{fig:8b_3090_accept}
\end{subfigure}
\hfill
\begin{subfigure}[b]{0.32\textwidth}
  \centering
  \begin{tikzpicture}
      \begin{axis}[
          xlabel={\scriptsize Draft tree size},
          ylabel={\scriptsize Verification/decoding},
          xmin=28, xmax=62,
          ymin=1, ymax=2.5,
          xtick={30,40,50,60},
          grid=major,
          width=\textwidth,
          height=0.75\textwidth,
          tick label style={font=\tiny},
          legend style={font=\tiny},
          legend pos=south west
      ]
      \addplot[color=StrongRed,mark=*] coordinates {(30, 1.1283068056447623)(40, 1.2947432421427232)(50, 1.3131251939926059)(60, 1.2619213258184565)}; 
      \addplot[color=MediumPurple,mark=*] coordinates {(30, 1.0944166487221878)(40, 1.118196549476688)(50, 1.1369944976308488)(60, 1.1584559610492287)}; 
      \addplot[color=DeepBlue,mark=*] coordinates {(30, 1.4749658335977536)(40, 1.9113201348167506)(50, 2.3347567876836055)(60, 2.370397279679798)}; 
      \addplot[color=EmeraldGreen,mark=*] coordinates {(30, 1.173119068710008)(40, 1.279413836850303)(50, 1.4465797928899657)(60, 1.4858585880325252)}; 
      \addplot[color=LightGreen, mark=*] coordinates {(30, 1.1774342173245538)(40, 1.344132070691125)(50, 1.5227973355496287)(60, 1.5636654100852125)};  
      \end{axis}
  \end{tikzpicture}
  \caption{Verification Ratio (8B, A100)}
  \label{fig:8b_3090_verification_ratio}
\end{subfigure}
\hfill
\begin{subfigure}[b]{0.32\textwidth}
  \centering
  \begin{tikzpicture}
      \begin{axis}[
          xlabel={\scriptsize Draft tree size},
          ylabel={\scriptsize Speedup},
          xmin=28, xmax=62,
          ymin=2, ymax=4.5,
          xtick={30,40,50,60},
          ytick={2, 2.5, 3, 3.5, 4, 4.5},
          grid=major,
          width=\textwidth,
          height=0.75\textwidth,
          tick label style={font=\tiny},
      ]
      \addplot[color=StrongRed,mark=*] coordinates {(30, 2.287007969405264)(40, 2.1601195989505224)(50, 2.14762543933708)(60, 2.199199431692463)}; 
      \addplot[color=MediumPurple,mark=*] coordinates {(30, 3.450918674809839)(40, 3.3949056187492284)(50, 3.275322155659007)(60, 3.105182713020926)}; 
      \addplot[color=DeepBlue,mark=*] coordinates {(30, 3.626918199646199)(40, 3.0847421830773336)(50, 2.677420169084853)(60, 2.569033514965405)}; 
      \addplot[color=EmeraldGreen,mark=*] coordinates {(30, 4.357687907142945)(40, 4.021861061200652)(50, 3.56362856452386)(60, 3.3415109640287426)}; 
      \addplot[color=LightGreen, mark=*] coordinates {(30, 4.2019060042777205)(40, 3.7747299023820076)(50, 3.383468553411035)(60, 3.1667240181690137)};  
      \end{axis}
  \end{tikzpicture}
  \caption{Speedup (8B, 3090)}
  \label{fig:8b_3090_speedup}
\end{subfigure}

\caption{Comparison of average accepted length, verification-to-decoding ratio, and speedup for various quantization precisions (FP16, W8A8, W4A16, W4A8-QQQ, W4A8-QQQ-g128) on Llama-3-8B with EAGLE-2, evaluated on A100 and RTX 3090. Panels (a–c) show A100 results; (d–f) show RTX 3090 results.}
\label{fig:8B_eagle_analysis}
\vspace{-5pt}
\end{figure*}
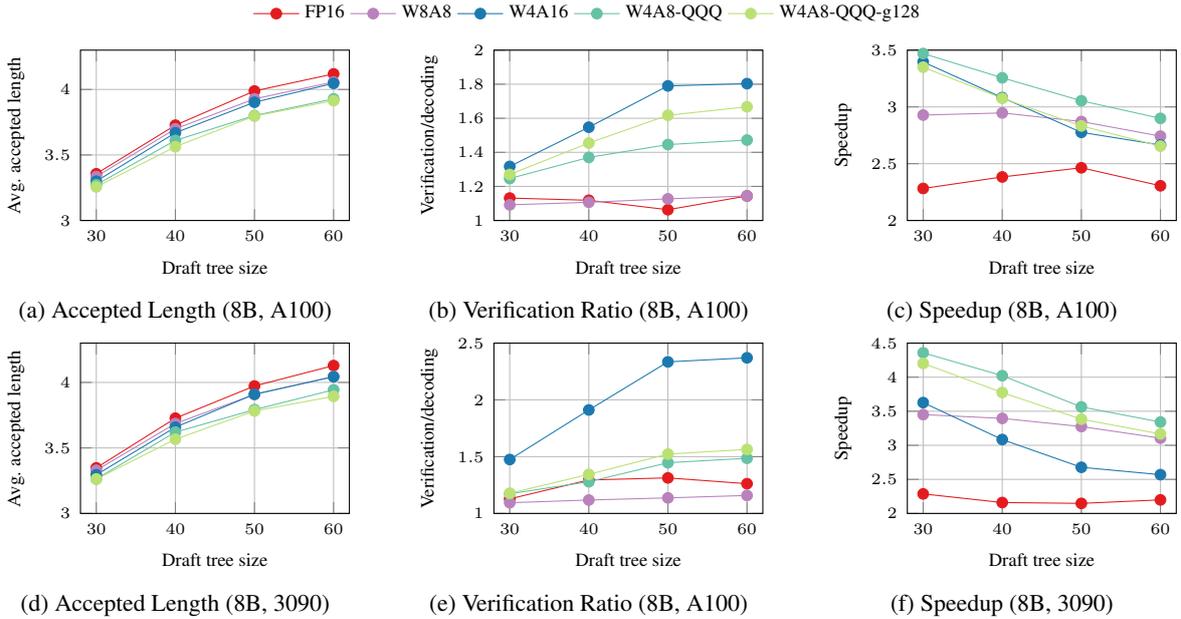

\subsection{Experimental Observation}
\label{subsec:comparison_speedup}

Figure~\ref{fig:diff_quant_ea_speedup} presents the speedup performance of speculative decoding EAGLE-2, various quantization methods, and their integration. It also includes the relative speedup improvement contributed by EAGLE-2 while integrating.

\textbf{EAGLE-2 vs. quantization.}
We compare the speedup achieved by applying EAGLE-2 to FP16 models with the speedup obtained through various quantization methods across different hardware platforms.
EAGLE‑2 achieves a higher speedup on the high-performance device A100 due to its computation-intensive design as shown in Figure~\ref{fig:diff_quant_ea_speedup}(a), while quantization of 4-bit weights (W4A16 and W4A8) yields higher gains on the customer-grad GPU RTX 3090 for its high reduction of memory demands demonstrated in Figure~\ref{fig:diff_quant_ea_speedup}(b).
Notably, EAGLE-2 is a lossless acceleration method, while quantization may introduce performance degradation, though W4A16 is considered to be nearly lossless.

\textbf{Integration and compatibility.}
The dashed bars in Figure~\ref{fig:diff_quant_ea_speedup} show that integrating EAGLE-2 with quantization yields additional speedup compared to using either technique alone, except when EAGLE-2 is applied to W4A16 8B model on the RTX 3090.
To evaluate the compatibility between EAGLE-2 and different quantization precisions, we present the relative speedup brought by EAGLE-2 across models with different precisions in Figure~\ref{fig:diff_quant_ea_speedup}.
We observe that, for models with 4-bit weight quantization (W4A16 and W4A8), the relative speedup provided by EAGLE-2 drops significantly compared to FP16 and W8A8, indicating lower compatibility, with W4A16 exhibiting the lowest compatibility.
This limited compatibility suggests a potential conflict between 4-bit weight quantization and the EAGLE-2 method.
A plausible hypothesis is that \textit{the increased computational overhead from EAGLE-2 diminishes the memory benefits of 4-bit weight optimization, resulting in limited speedup.}

\subsection{Factors behind Integration Conflict}
\label{subsec:eagle_analysis}

As indicated in Equation~\ref{eq:speedup}, the speedup of speculative decoding depends on three key factors: the average accepted length $\tau(n, d)$, the draft to target decoding time ratio $T_d/T_t$, and the target verification to decoding time ratio $T_v(n)/T_t$.
To further understand the conflict underlying the limited compatibility between EAGLE-2 and 4-bit weight quantized models, we analyze the impact of three factors on overall speedup when applying EAGLE-2 to various quantization schemes.

We perform experiments by varying the draft tree size $n$ and corresponding draft forward passes $d$, the only variables in Equation~\ref{eq:speedup} that are strongly correlated with the three factors.
For Llama-3-8B and Llama-3-70B, we use $n \in \{30, 40, 50, 60\}$ and $n \in \{24, 32, 40, 48\}$ respectively, with corresponding draft forward passes $d \in \{3, 4, 5, 6\}$. 
For W4A8, we use the symmetric quantization method QQQ, as it offers higher decoding speedup. 

\begin{figure*}[t]
  \centering
  \begin{tikzpicture}
    \begin{axis}[
      hide axis,
      xmin=0,xmax=1,ymin=0,ymax=1,
      legend style={
        draw=black,     
        font=\scriptsize,
        legend columns=-1,
        at={(0.5,1)},    
        anchor=south,
        draw=none,        
        fill=none,        
      }
    ]
    \addlegendimage{MediumPurple,mark=*}
    \addlegendimage{DeepBlue,mark=*}
    \addlegendimage{EmeraldGreen,mark=*}
    \addlegendimage{LightGreen,mark=*}
    \legend{ W8A8, W4A16, W4A8-QQQ, W4A8-QQQ-g128}
    \end{axis}
  \end{tikzpicture}

\begin{subfigure}[b]{0.32\textwidth}
  \centering
  \begin{tikzpicture}
      \begin{axis}[
          xlabel={\scriptsize Draft tree size},
          ylabel={\scriptsize Avg. accepted length},
          xmin=22, xmax=50,
          ymin=3, ymax=4.3,
          xtick={24,32,40,48},
          grid=major,
          width=\textwidth,
          height=0.75\textwidth,
          tick label style={font=\tiny},
          legend style={font=\tiny},
          legend pos=south west
      ]
      \addplot[color=MediumPurple,mark=*] coordinates {(24, 3.261952615871855)(32, 3.62648420205272)(40, 3.7849396773756268)(48, 3.894793536804309)}; 
      \addplot[color=DeepBlue,mark=*] coordinates {(24, 3.28285417811145)(32, 3.6302532327586206)(40, 3.818927132659551)(48, 3.9262319052493053)};
      \addplot[color=EmeraldGreen,mark=*] coordinates {(24, 3.2039601554907677)(32, 3.4952228235451326)(40, 3.6638315672058)(48, 3.749591067491643)}; 
      \addplot[color=LightGreen, mark=*] coordinates {(24, 3.2426872193227334)(32, 3.588569518716577)(40, 3.7754210415051794)(48, 3.8767813444652366)}; 
      \end{axis}
  \end{tikzpicture}
  \caption{Accepted Length (70B, A100)}
  \label{fig:70b_a100_accept}
\end{subfigure}
\hfill
\begin{subfigure}[b]{0.32\textwidth}
  \centering
  \begin{tikzpicture}
      \begin{axis}[
          xlabel={\scriptsize Draft tree size},
          ylabel={\scriptsize Verification/decoding},
          xmin=22, xmax=50,
          ymin=1, ymax=1.8,
          xtick={24, 32, 40, 48},
          grid=major,
          width=\textwidth,
          height=0.75\textwidth,
          tick label style={font=\tiny},
          legend style={font=\tiny},
          legend pos=south west
      ]
      \addplot[color=MediumPurple,mark=*] coordinates {(24, 1.0849497474905678)(32, 1.11283792866236)(40, 1.141003921909717)(48, 1.168492643682216)}; 
      \addplot[color=DeepBlue,mark=*] coordinates {(24, 1.2615674528882275)(32, 1.286964181422071)(40, 1.5551623213274772)(48, 1.579999091277826)};
      \addplot[color=EmeraldGreen,mark=*] coordinates {(24, 1.325802029474064)(32, 1.3518514788740275)(40, 1.480845249127147)(48, 1.5201325631002187)}; 
      \addplot[color=LightGreen, mark=*] coordinates {(24, 1.3536119250737502)(32, 1.376189601207407)(40, 1.59825362046829)(48, 1.629632924337803)};  
      \end{axis}
  \end{tikzpicture}
  \caption{Verification Ratio (70B, A100)}
  \label{fig:70b_a100_verification_ratio}
\end{subfigure}
\hfill
\begin{subfigure}[b]{0.32\textwidth}
  \centering
  \begin{tikzpicture}
      \begin{axis}[
          xlabel={\scriptsize Draft tree size},
          ylabel={\scriptsize Speedup},
          xmin=22, xmax=50,
          ymin=2.5, ymax=4.5,
          xtick={24, 32, 40, 48},
          ytick={2.5, 3, 3.5, 4.0, 4.5},
          grid=major,
          width=\textwidth,
          height=0.75\textwidth,
          tick label style={font=\tiny},
          legend style={font=\tiny, at={(2,1.05)}, anchor=south, legend columns=-1},
          % legend pos=south west
      ]
      \addplot[color=MediumPurple,mark=*] coordinates {(24, 2.638540149322329)(32, 2.7553200829298112)(40, 2.7582159648565416)(48, 2.695243434661454)}; 
      \addplot[color=DeepBlue,mark=*] coordinates {(24, 3.7309859968630485)(32, 3.8603426662193496)(40, 3.3746903830484514)(48, 3.297596779230136)}; 
      \addplot[color=EmeraldGreen,mark=*] coordinates {(24, 3.565933904250844)(32, 3.6375977941130113)(40, 3.4084347753863953)(48, 3.2811463167988655)}; 
      \addplot[color=LightGreen, mark=*] coordinates {(24, 3.386641764842042)(32, 3.5156793583284722)(40, 3.186623806760057)(48, 3.0923810722867144)};  
      \end{axis}
  \end{tikzpicture}
  \caption{Speedup (70B, A100)}
  \label{fig:70b_a100_speedup}
\end{subfigure}
  \caption{Comparison of average accepted length, verification-to-decoding ratio, and speedup for various quantization precisions (W8A8, W4A16, W4A8-QQQ, W4A8-QQQ-g128) on Llama-3-70B with EAGLE-2 on A100.}
  \vspace{-5pt}
  \label{fig:70B_eagle_analysis}
\end{figure*}
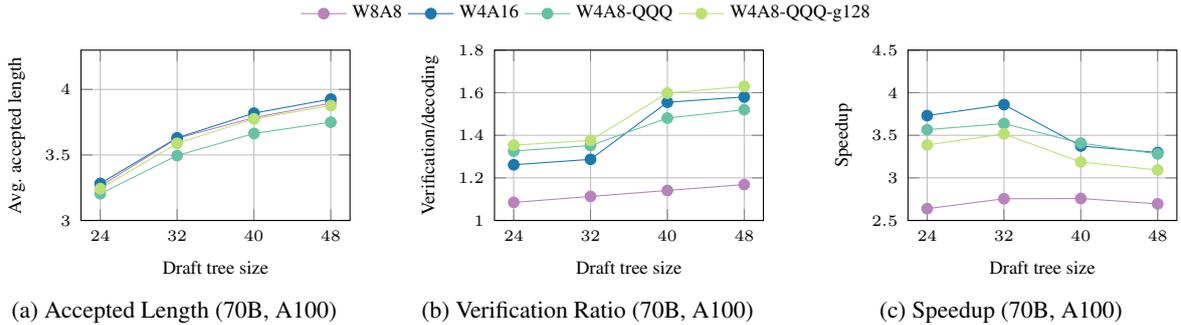

Experimental results are shown in Figures~\ref{fig:8B_eagle_analysis} and~\ref{fig:70B_eagle_analysis}, with speedup in Figures~\ref{fig:8b_a100_speedup}, \ref{fig:8b_3090_speedup}, and~\ref{fig:70b_a100_speedup}.
We find that the size of drafts significantly affects the overall speedup, with \textbf{fewer drafts yielding a higher speedup in 4-bit weight models.}

\textbf{\textit{Quantization has minimal impact on average accepted length.}}
For different quantization precisions in Figure~\ref{fig:8b_a100_accept},~\ref{fig:8b_3090_accept}, and~\ref{fig:70b_a100_accept}, the decrease in the average accepted length $\tau$ caused by quantization is minimal compared to FP16, with W4A16 and W8A8 exhibiting nearly no degradation.
However, although the average accepted length $\tau$ increases with draft tree size, this increase does not result in improved speedup for 4-bit weight models.
It suggests that changes in average accepted length are not the primary factor affecting the integrated speedup for 4-bit weight models.

\textbf{\textit{Higher draft-to-target decoding time ratio from quantization partly contributes to the higher speedup of fewer drafts.}}
The ratio of draft to target decoding time $T_d/T_t$ is another factor affecting the speedup. 
The decoding speed of the 4-bit weight quantized model is improved compared to the FP16 model as shown in Figure~\ref{fig:diff_quant_ea_speedup}, which in turn increases the ratio of draft to target decoding time $T_d/T_t$. 
According to ~\citet{leviathan2023fast}, if the acceptance rate remains almost unchanged while this ratio increases, the optimal speedup is achieved with fewer draft forward passes.
This indicates that the increased draft-to-target decoding ratio is one reason why fewer drafts yield higher speedup.

\textbf{\textit{Verification-to-decoding time ratio of 4-bit weight models favors fewer drafts and undermines EAGLE-2 compatibility.}}
As shown in Figures~\ref{fig:8b_a100_verification_ratio},~\ref{fig:8b_3090_verification_ratio}, and~\ref{fig:70b_a100_verification_ratio}, 4-bit weight quantized models exhibit a significantly greater increase trend in the verification-to-decoding time ratio $T_v(n)/T_t$ with draft tree size compared to FP16 and W8A8 models.
For example, the 8B model on the A100 with a draft tree size of 60 shows a ratio below 1.2 for FP16 and W8A8, while W4A16 reaches 1.8, significantly exceeding the ideal value of 1.
As a consequence, the increase in the verification-to-decoding ratio for 4-bit quantized models with growing draft tree size drives the decline in integrated speedup, despite gains from longer average accepted length.
This clearly indicates the incompatibility between 4-bit weight models and the EAGLE-2 method, where the heavy computation time $T_v(n)$ required during draft verification undermines the memory efficiency gained by the 4-bit weight quantization.

\begin{figure}[t]
  \centering
  \begin{tikzpicture}
    \begin{axis}[
        name=main,
        width=0.95\columnwidth,
        height=0.20\textwidth,
        ybar=2pt,
        axis lines*=left,
        ymin=2, ymax=4,
        ylabel={Speedup},
        xmin=-0.5, xmax=2.5,
        xtick={0,1,2},
        xticklabels={FP16(8B), W4A16(8B), W4A16(70B)},
        xticklabel style={font=\scriptsize, yshift=-2pt},
        ylabel style={font=\scriptsize},
        yticklabel style={font=\scriptsize},
        ymajorgrids=true,
        grid style=dashed,
        bar width=10pt,
        nodes near coords,
        nodes near coords style={
          font=\tiny,
          inner sep=1pt,
          /pgf/number format/.cd,
          fixed,
          precision=1
        },
        legend style={ 
          font=\scriptsize,  
          at={(0.5,1.07)},
          anchor=south,
          legend columns=7,
          /tikz/every even column/.append style={column sep=0.2cm},
          draw=none,         
          fill=none,        
          inner sep=1pt
        },
        legend image code/.code={
          \draw[#1,fill=#1] (0cm,-0.1cm) rectangle (0.24cm,0.08cm);
        },
    ]
    \addplot[bar shift=0, fill=myLavenderBlue] coordinates {(-0.24,2.3054998665481254) (0.76,2.6648778328104377) (1.64, 3.297596779230136)}; 
    \addplot[bar shift=0, fill=myLilac] coordinates {(0.0, 2.1457465619043705) (1.0, 2.9131105656764564) (1.88, 3.579979329650579)}; 
    \addplot[bar shift=0, fill=myPeriwinkle] coordinates {(0.24, 2.281886166758471) (1.24, 3.396056945230414) (2.12, 3.7309859968630485)}; 

    \addplot[bar shift=0, fill=myIceBlue] coordinates {(2.36, 3.8921445746950494 )}; 
    
    \legend{{EG-2 (6,full)}, {EG-2 (6,half)}, {EG-2 (3,half)}, {SP (6)}}
    \end{axis}
  
    \end{tikzpicture}
    
    \caption{Speedup comparison of different EAGLE-2 configurations and vanilla speculative decoding on Llama-3 models. EG-2(6/3, full/half) uses 6 or 3 draft passes with full (60, 48) or half (30, 24) tree sizes; SP(6) denotes vanilla speculative decoding with 6 draft passes.} 
    \label{fig:computation_confirm}
    \vspace{-5pt}
\end{figure}
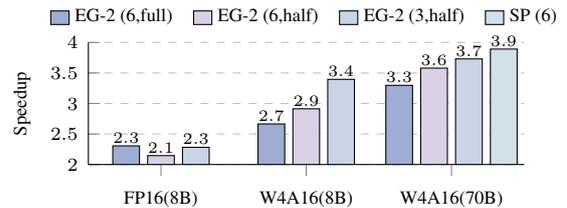

To confirm that the verification-to-decoding ratio and draft-to-decoding ratio contribute to the negative correlation between speedup and draft-tree size in 4-bit quantized models, we compare the speedup of FP16 and W4A16 models with EAGLE-2 under three draft configurations : (1) full-sized tree with 6 forward passes, (2) half-sized tree with 6 forward passes, and (3) half-sized tree with 3 forward passes.
Figure~\ref{fig:computation_confirm} shows that reducing tree size decreases speedup for FP16 but improves it for W4A16, indicating that smaller trees better preserve memory efficiency under W4A16's higher verification-to-decoding ratio.
Further W4A16 speedup from fewer draft forward passes also confirms the impact of the higher draft-to-decoding ratio, which is more pronounced in 8B models, since the gap between draft and target model sizes is smaller than that in 70B models.

In addition, to demonstrate that the increased verification-to-decoding ratio primarily causes the conflict between 4-bit quantization and EAGLE-2 in memory optimization, we compare two methods on a W4A16 70B model: EAGLE-2 and the vanilla speculative decoding. 
For the latter, we employ a W4A16 8B draft model to generate sequence-style drafts.
As shown in Figure~\ref{fig:computation_confirm}, despite the increased draft overhead, the vanilla speculative decoding approach outperforms EAGLE-2 due to the low computational cost of its sequential draft verification.
This result highlights that the heavy computational cost of tree-style draft verification in EAGLE-2 undermines the memory access advantages of 4-bit weight quantization, explaining the conflict underlying their limited compatibility.

\section{Hierarchical Framework}
For near-lossless W4A16 models, EAGLE-2 is mainly bottlenecked by tree-style draft verification, while vanilla speculative decoding is limited by drafting overhead.
To overcome the limitations of both methods, we propose a hierarchical speculative decoding framework.
We also evaluate the hierarchical framework across multiple tasks.
\subsection{Methodology}
Building upon the EAGLE-2 framework, we introduce a hierarchical design for W4A16 models by introducing a small intermediate model between the draft and target stages.
We adjust the original EAGLE-2 to perform tree-style speculation for the small intermediate model, whose outputs subsequently serve as the draft inputs for the final target model.
This hierarchical framework mainly consists of two levels: the compute-intensive drafting stage and the memory-efficient verification stage.
Through the small model as a bridge, it achieves an effective combination of efficient tree-style drafting and sequential draft verification.

\textbf{Computation-intensive drafting stage.}
This stage accelerates the small model drafting using EAGLE-2. 
In each draft iteration, the lightweight EAGLE-2 draft model generates an initial draft of length $d_1$, and the small model performs multi-token drafting through verification. 
This process repeats until the draft length of the small model exceeds draft length $d$.
The small model undertakes the computation-intensive tree-based verification, converting the tree-based drafts into sequential drafts, thus enabling high-speed drafting.

\textbf{Memory-efficient verification stage.}
In this verification stage, the target model verifies the sequential draft with length $d$ from the smaller model to generate multiple tokens in one forward pass. 
Compared to directly applying tree-based draft verification, sequential draft verification enables further memory access optimization while fully retaining the memory advantages of 4-bit weight models, achieving orthogonality between speculative decoding and 4-bit weight quantization.

\begin{table*}[t]
  \centering
  \setlength{\tabcolsep}{4pt}
  \scalebox{0.75}{
    \begin{tabular}{lc|cccccccccccc|cc}
      \toprule
      \multirow{2}{*}{\textbf{Method}} 
    & \multirow{2}{*}{$d$}
      & \multicolumn{2}{c}{\textbf{MT.}}
      & \multicolumn{2}{c}{\textbf{Conv.}}
      & \multicolumn{2}{c}{\textbf{RAG}}
      & \multicolumn{2}{c}{\textbf{Math}}
      & \multicolumn{2}{c}{\textbf{QA}}
      & \multicolumn{2}{c}{\textbf{Summ.}}
      & \multicolumn{2}{|c}{\textbf{Avg.}} \\
      \cmidrule(lr){3-4}\cmidrule(lr){5-6}\cmidrule(lr){7-8}\cmidrule(lr){9-10}\cmidrule(lr){11-12}\cmidrule(lr){13-14}\cmidrule(l){15-16}
& 
& $\tau$ & \textbf{Tok/s}
& $\tau$ & \textbf{Tok/s}
& $\tau$ & \textbf{Tok/s}
& $\tau$ & \textbf{Tok/s}
& $\tau$ & \textbf{Tok/s}
& $\tau$ & \textbf{Tok/s}
& $\tau$ & \textbf{Tok/s} \\
\midrule
Vanilla AR & -
& 1.00 & 35.86
& 1.00 & 35.08
& 1.00 & 35.02
& 1.00 & 35.49
& 1.00 & 35.66
& 1.00 & 34.59
& 1.00 & 35.28 (1.0 $\times$) \\
\midrule
\multirow{2}{*}{Vanilla SP}
& 6
& 4.45 & 78.54
& 4.71 & 79.77
& 4.81 & 86.55
& 5.57 & 92.07
& 4.28 & 73.64
& 4.66 & 75.40
& 4.72 & 81.33 (2.31$\times$) \\
& 7
& 4.74 & 77.75
& 5.10 & 80.24
& 5.19 & 87.36
& 6.12 & 95.57
& 4.55 & 73.06
& 5.00 & 74.77
& 5.09 & 81.46 (2.31$\times$) \\

\midrule
\multirow{4}{*}{EAGLE-2} 
& 3 
& 3.18 & 74.14
& 3.28 & 74.48
& 3.40 & 76.98
& 3.49 & 79.54
& 2.99 & 68.88
& 3.16 & 70.34
& 3.25 & 74.06 (2.10$\times$) \\
& 4
& 3.44 & 74.27
& 3.63 & 76.51
& 3.77 & 79.03
& 3.92 & 82.41
& 3.19 & 67.61
& 3.42 & 69.78
& 3.57 & 74.93 (2.12$\times$) \\
& 5
& 3.60 & 64.61
& 3.82 & 67.46
& 4.00 & 70.58
& 4.14 & 72.18
& 3.28 & 57.78
& 3.56 & 60.65
& 3.73 & 65.54 (1.86$\times$) \\
& 6
& 3.65 & 62.41
& 3.93 & 66.22
& 4.09 & 68.06
& 4.23 & 70.21
& 3.31 & 55.45
& 3.59 & 58.29
& 3.81 & 63.44 (1.80$\times$) \\
\midrule

\multirow{4}{*}{\textbf{HierSpec}}
& 6(3)
& 4.84 & \textbf{92.12}
& 5.30 & \textbf{98.23}
& 5.35 & 105.65
& 6.58 & 120.29
& 4.62 & \textbf{83.61}
& 5.16 & \textbf{89.21}
& 5.28 & \textbf{98.19 (2.78$\times$)} \\
& 6(4)
& 4.82 & 86.02
& 5.17 & 90.45
& 5.28 & 98.76
& 6.36 & 112.12
& 4.61 & 77.13
& 5.14 & 82.34
& 5.19 & 91.14 (2.58$\times$) \\
& 7(3)
& 4.99 & 91.06
& 5.47 & 97.79
& 5.56 & \textbf{106.09}
& 6.82 & \textbf{120.89}
& 4.80 & 83.39
& 5.37 & 88.76
& 5.46 & 98.00 (2.78$\times$) \\
& 7(4)
& 5.12 & 86.29
& 5.64 & 93.27
& 5.75 & 102.75
& 7.18 & 116.70
& 4.89 & 76.94
& 5.48 & 82.45
& 5.62 & 93.06 (2.64$\times$) \\

\bottomrule       
    \end{tabular}
  }
  \caption{Average accepted length $\tau$ and decoding speed \textbf{Tok/s} of different methods on W4A16 Llama-3-70B with different draft lengths $d$. The numbers in draft length parentheses (e.g., 3 in 6(3)) denote the draft length $d_1$ of EAGLE-2 for the small model in the first hierarchy. And the numbers in decoding speed parentheses (2.78$\times$) represent the speedup compared to the W4A16 vanilla auto-regressive approach.}
  \label{tab:hspec_experiment}
  \vspace{-5pt}
\end{table*}

\begin{table}[h]
\centering
\setlength{\tabcolsep}{4pt}
\scalebox{0.76}{
\begin{tabular}{l|ccccccc}
\toprule
Method     & MT.  & Conv. & RAG  & Math & QA   & Summ. & Avg.\\
\midrule
Vanilla SP & 60.9 & 71.6 & 91.8 & 63.3 & 57.3 & 95.3 & 73.4 \\
EAGLE-2    & 3.8  & 6.4  & 9.4  & 4.8  & 3.7  & 10.3  & 6.4 \\
HierSpec   & 63.7 & 75.1 & 94.1 & 65.3 & 59.9 & 100.2 & 76.4 \\
\bottomrule
\end{tabular}
}
\caption{Draft latency (ms) of different methods on W4A16 Llama-3-70B on A100.}
\label{tab:draft_latency_llama3}
\vspace{-5pt}
\end{table}

\subsection{Experimental Setup}
\label{sec:hier_setup}

\textbf{Dataset.}
We use SpecBench~\citep{xia-etal-2024-unlocking} as the evaluation dataset for our method, which includes six types of text generation tasks: machine translation (MT.), multi-turn conversation (Conv.), retrieval-augmented generation (RAG), arithmetic reasoning (Math), question answering (QA), and document summarization (Summ.).

\textbf{Models and Evaluation.} 
We adopted W4A16 Llama-3-70B as the target model.
We evaluate our hierarchical speculative decoding framework (HierSpec) and three baselines: vanilla auto-regressive decoding (Vanilla AR), vanilla speculative decoding (Vanilla SP), and EAGLE-2 method.
W4A16 Llama-3-8B serves as the small intermediate model in HierSpec and the draft model in Vanilla SP.
Average acceptance length $\tau$ and decoding speed (tokens/s) are reported.
Given our hierarchical design, we also report the draft latency (the draft model's prefilling time), excluded from decoding time.
For a systematic comparison, we vary the draft length $d$: $\{6, 7\}$ for Vanilla SP, and $d \in \{3, 4, 5, 6\}$ with corresponding tree sizes $\{24, 32, 40, 48\}$ for EAGLE-2, following Section~\ref{subsec:eagle_analysis}. 
For HierSpec, the first-level draft length $d_1 \in \{3, 4\}$ matches the optimal EAGLE-2 setting for 8B, and the second-level $d \in \{6, 7\}$ follows Vanilla SP.
All experiments are conducted on a single NVIDIA 80GB A100 GPU.

\begin{figure}[h]
  \centering
  \begin{tikzpicture}
    \begin{axis}[
        name=main,
        width=0.9\columnwidth,
        height=0.20\textwidth,
        ybar=2pt,
        axis lines*=left,
        ymin=0, ymax=10,
        ylabel={Draft Time (ms)},
        ytick={0, 2, 4, 6, 8, 10},
        xmin=-0.5, xmax=2.5,
        xtick={0,1,2},
        xticklabels={EAGLE-2, Vanilla SP, HierSpec},
        xticklabel style={font=\scriptsize, yshift=-2pt},
        ylabel style={font=\scriptsize},
        yticklabel style={font=\scriptsize},
        ymajorgrids=true,
        grid style=dashed,
        bar width=10pt,
        nodes near coords,
        nodes near coords style={
          font=\tiny,
          inner sep=1pt,
          /pgf/number format/.cd,
          fixed,
          precision=1
        },
        legend style={ 
          font=\scriptsize,  
          at={(0.25,1.07)},
          anchor=south,
          legend columns=3,
          /tikz/every even column/.append style={column sep=0.2cm},
          draw=none,         
          fill=none,         
          inner sep=1pt
        },
        legend image code/.code={
          \draw[#1,fill=#1] (0cm,-0.1cm) rectangle (0.24cm,0.08cm);
        },
    ]

    \addplot[bar shift=0, fill=myLavenderBlue] coordinates {(-0.14,2.748) (0.86, 5.174) (1.86, 3.81)};
    \legend{{Drafting time}}

    \end{axis}

        \begin{axis}[
        name=main,
        width=0.9\columnwidth,
        height=0.20\textwidth,
        ybar=2pt,
        axis y line*=right,
        axis x line=none,
        ymin=0, ymax=50,
        ylabel={Verif.\ Time (ms)},
        ytick={0, 10, 20, 30, 40, 50},
        xmin=-0.5, xmax=2.5,
        ylabel style={font=\scriptsize},
        yticklabel style={font=\scriptsize},
        bar width=10pt,
        nodes near coords,
        nodes near coords style={
          font=\tiny,
          inner sep=1pt,
          /pgf/number format/.cd,
          fixed,
          precision=1,
          zerofill,
        },
        legend style={ 
          font=\scriptsize,  
          at={(0.75,1.07)},
          anchor=south,
          legend columns=3,
          /tikz/every even column/.append style={column sep=0.2cm},
          draw=none,         
          fill=none,         
          inner sep=1pt
        },
        legend image code/.code={
          \draw[#1,fill=#1] (0cm,-0.1cm) rectangle (0.24cm,0.08cm);
        },
    ]

    \addplot[bar shift=0, fill=myLilac] coordinates {(0.14, 45.274) (1.14, 29.688) (2.14, 30.03)}; 
    \legend{ {Verfication time}}

    \end{axis}
  
    \end{tikzpicture}
    
    \caption{Comparison of drafting time (per draft length) and verification time of three speculative decoding methods applied on W4A16 Llama-3-70B on A100.}
    \label{fig:draft_and_verification_time}
    \vspace{-5pt}
\end{figure}
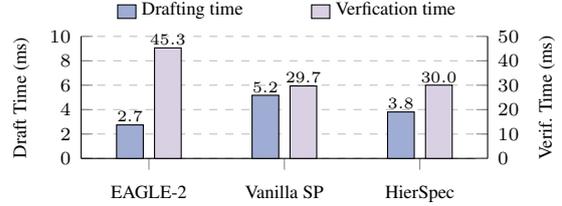

\subsection{Main Results}
Table~\ref{tab:hspec_experiment} compares the average acceptance length $\tau$ and decoding speed (tokens/s) of different methods on W4A16 Llama-3-70B with different draft lengths. 
Experiment results demonstrate that our proposed HierSpec achieves the highest decoding speed on average and also consistently outperforms baselines across various tasks.
Specifically, HierSpec with $d_{1}=3$ and $d=6$ outperforms the other configurations, achieving an average decoding speed 98.19 tokens/s, which is 1.31$\times$ over the best of EAGLE-2 and 1.21$\times$ over the best of Vanilla SP. 
In addition, the average acceptance length of Hierspec shows some increase compared to Vanilla SP, as the tree-level drafting in the first hierarchical level generates additional tokens beyond the draft length $d$.
Table~\ref{tab:draft_latency_llama3} reports the draft latency (ms) of different methods.
Although our hierarchical framework has a larger prefill overhead, this cost is amortized over decoding steps, making it more suitable for long text generation.

Figure~\ref{fig:draft_and_verification_time} compares the drafting time per draft length and verification time per iteration across speculative decoding methods at $d=6$.
Vanilla SP exhibits a drafting bottleneck with 1.93$\times$ longer drafting time than EAGLE-2, while EAGLE-2 faces a verification bottleneck with 1.53$\times$ longer verification time than Vanilla SP. 
Our HierSpec approaches EAGLE-2 in drafting time and matches Vanilla SP in verification time.
It combines efficient drafting with memory-efficient verification, leading to further speedup in the integration of speculative decoding with 4-bit weight quantization.

\subsection{Integration with EAGLE-3}

We further assess HierSpec by integrating with the EAGLE-3~\citep{li2025eagle} method.
Given that the training details of EAGLE-3 remain unpublished, we can only experiment with the publicly available checkpoints.
For the 70B model size, only Llama-3.3-70B has an available EAGLE-3 checkpoint, which we adopt as our baseline.
For our HierSpec method, due to the absence of an 8B model in the Llama-3.3 series, we can only adopt an 8B model from the Llama-3.1 series as the intermediate model. 
Even under this unfair configuration, HierSpec still achieves a further speedup over EAGLE-3, as shown in Appendix~\ref{sec:appendix}.

\section{Related Work}

This section introduces the related work on accelerating LLM inference using quantization and speculative decoding.

\subsection{Quantization of LLMs}
Quantization reduces LLM memory footprint and accelerates inference.
\textit{Weight-only quantization} only quantizes the model weights and is highly effective for memory-bound LLM inference.
GPTQ~\citep{frantar2022gptq} uses second-order information to minimize rounding errors, while AWQ~\cite{lin2024awq} and SqueezeLLM~\cite{kim2024squeezellm} prioritize important weights.
QuIP\#~\citep{tseng2024quip} applies pre-quantization transformations.
\textit{Weight-activation quantization} further reduces computation by quantizing both weights and activations.
To alleviate the impact of activation outliers, LLM.int8()~\citep{dettmers2022gpt3} performs mixed-precision decomposition, SmoothQuant~\citep{xiao2023smoothquant} and OmniQuant~\citep{shao2024omniquant} adopt per-channel scaling transformation.
Further, QuaRot~\cite{ashkboos2024quarot} and FlatQuant~\citep{sun2024flatquant} apply Hadamard transformation and Affine Transformation, respectively.
\textit{KV cache quantization}~\cite{liu2024kivi, hooper2024kvquant} quantizes key and value caches to mitigate the KV bottleneck in the long-context inference. 
However, long-context inference is not included in our work.

\subsection{Speculative Decoding}

Speculative decoding~\cite{leviathan2023fast,chen2023accelerating} accelerates LLM inference by drafting multiple tokens and verifying them in parallel.
One line of work uses external and lightweight components for draft generation to enable low-cost drafting.
Medusa~\citep{cai2024medusa} and EAGLE series works ~\cite{li2024eagle1,li2024eagle,li2025eagle} employ an LM head and a single Transformer layer as draft models with tree-style drafting, respectively.
EAGLE-3~\cite{li2025eagle} further achieves state-of-the-art performance via Training-Time Test.

Another line of work, self-speculative decoding, shares the model architecture between draft and target models for better alignment.
Triforce~\citep{sun2024triforce} and MagicDec~\citep{sadhukhan2025magicdec} draft with sparse KV-cache to mitigate KV bottleneck.
Recently, several studies also integrated quantization into self-speculative frameworks.
QSpec~\citep{zhao2024qspec} accelerates batch inference for 4-bit weight-only models using shared weights and 4-bit activations, but fails in single-batch settings.
QuantSpec~\citep{tiwari2025quantspec} improves long-context efficiency using 4-bit weights and KV caches, though limited gains on short contexts.
ML-SpecQD~\citep{georganas2025ml} adopts the 4-bit target model as the draft model and further introduces a tiny 4-bit model for multi-level speculative decoding, but the FP16 target model limits deployment.

However, our work focuses on integrating speculative decoding with a lightweight draft model into quantized target models for further acceleration.

\section{Conclusion}
In this work, we systematically study the compatibility of speculative decoding and quantization when applied jointly to LLMs under various precisions and speculative decoding configurations. 
Our study reveals that the substantial computation overhead from the tree-style verification of speculative decoding undermines the memory bandwidth benefits of 4-bit weight quantization.
Motivated by this finding, we propose a hierarchical speculative decoding framework for W4A16 models, leveraging a small model as a bridge to enable both efficient drafting and memory-efficient verification. 
Experimental results show that our method achieves a 2.78$\times$ speedup over auto-regressive decoding and a 1.31$\times$ speedup over the EAGLE-2 approach, enhancing the compatibility of these two techniques. 

\section*{Limitations}
Our current research focuses primarily on weight-only quantization and weight-activation quantization across several common tasks, while lacking assessments under some conditions, such as long-context tasks with KV cache quantization. Nevertheless, our study remains systematic and comprehensive, providing a more effective speculative decoding framework based on our findings. In the future, we will explore the integration of these two techniques in more challenging tasks.

\bibliography{custom}

\begin{thebibliography}{38}
\providecommand{\natexlab}[1]{#1}

\bibitem[{Ashkboos et~al.(2024)Ashkboos, Mohtashami, Croci, Li, Cameron, Jaggi, Alistarh, Hoefler, and Hensman}]{ashkboos2024quarot}
Saleh Ashkboos, Amirkeivan Mohtashami, Maximilian Croci, Bo~Li, Pashmina Cameron, Martin Jaggi, Dan Alistarh, Torsten Hoefler, and James Hensman. 2024.
\newblock Quarot: Outlier-free 4-bit inference in rotated llms.
\newblock In \emph{Proceedings of NeurIPS}, pages 100213--100240.

\bibitem[{Brown et~al.(2020)Brown, Mann, Ryder, Subbiah, Kaplan, Dhariwal, Neelakantan, Shyam, Sastry, Askell et~al.}]{brown2020language}
Tom Brown, Benjamin Mann, Nick Ryder, Melanie Subbiah, Jared~D Kaplan, Prafulla Dhariwal, Arvind Neelakantan, Pranav Shyam, Girish Sastry, Amanda Askell, and 1 others. 2020.
\newblock Language models are few-shot learners.
\newblock In \emph{Proceedings of NeurIPS}, pages 1877--1901.

\bibitem[{Cai et~al.(2024)Cai, Li, Geng, Peng, Lee, Chen, and Dao}]{cai2024medusa}
Tianle Cai, Yuhong Li, Zhengyang Geng, Hongwu Peng, Jason~D. Lee, Deming Chen, and Tri Dao. 2024.
\newblock Medusa: Simple {LLM} inference acceleration framework with multiple decoding heads.
\newblock In \emph{Proceedings of ICML}, pages 5209--5235.

\bibitem[{Chen et~al.(2023)Chen, Borgeaud, Irving, Lespiau, Sifre, and Jumper}]{chen2023accelerating}
Charlie Chen, Sebastian Borgeaud, Geoffrey Irving, Jean-Baptiste Lespiau, Laurent Sifre, and John Jumper. 2023.
\newblock Accelerating large language model decoding with speculative sampling.
\newblock \emph{arXiv preprint arXiv:2302.01318}.

\bibitem[{Chen et~al.(2021)Chen, Tworek, Jun, Yuan, Pinto, Kaplan, Edwards, Burda, Joseph, Brockman et~al.}]{chen2021evaluating}
Mark Chen, Jerry Tworek, Heewoo Jun, Qiming Yuan, Henrique Ponde De~Oliveira Pinto, Jared Kaplan, Harri Edwards, Yuri Burda, Nicholas Joseph, Greg Brockman, and 1 others. 2021.
\newblock Evaluating large language models trained on code.
\newblock \emph{arXiv preprint arXiv:2107.03374}.

\bibitem[{Cobbe et~al.(2021)Cobbe, Kosaraju, Bavarian, Chen, Jun, Kaiser, Plappert, Tworek, Hilton, Nakano et~al.}]{cobbe2021training}
Karl Cobbe, Vineet Kosaraju, Mohammad Bavarian, Mark Chen, Heewoo Jun, Lukasz Kaiser, Matthias Plappert, Jerry Tworek, Jacob Hilton, Reiichiro Nakano, and 1 others. 2021.
\newblock Training verifiers to solve math word problems.
\newblock \emph{arXiv preprint arXiv:2110.14168}.

\bibitem[{Dettmers et~al.(2022)Dettmers, Lewis, Belkada, and Zettlemoyer}]{dettmers2022gpt3}
Tim Dettmers, Mike Lewis, Younes Belkada, and Luke Zettlemoyer. 2022.
\newblock Gpt3. int8 (): 8-bit matrix multiplication for transformers at scale.
\newblock In \emph{Proceedings of NeurIPS}, pages 30318--30332.

\bibitem[{Frantar et~al.(2022)Frantar, Ashkboos, Hoefler, and Alistarh}]{frantar2022gptq}
Elias Frantar, Saleh Ashkboos, Torsten Hoefler, and Dan Alistarh. 2022.
\newblock Gptq: Accurate post-training quantization for generative pre-trained transformers.
\newblock \emph{arXiv preprint arXiv:2210.17323}.

\bibitem[{Frantar et~al.(2025)Frantar, Castro, Chen, Hoefler, and Alistarh}]{frantar2025marlin}
Elias Frantar, Roberto~L Castro, Jiale Chen, Torsten Hoefler, and Dan Alistarh. 2025.
\newblock Marlin: Mixed-precision auto-regressive parallel inference on large language models.
\newblock In \emph{Proceedings of PPoPP}, pages 239--251.

\bibitem[{Gao et~al.(2020)Gao, Biderman, Black, Golding, Hoppe, Foster, Phang, He, Thite, Nabeshima et~al.}]{gao2020pile}
Leo Gao, Stella Biderman, Sid Black, Laurence Golding, Travis Hoppe, Charles Foster, Jason Phang, Horace He, Anish Thite, Noa Nabeshima, and 1 others. 2020.
\newblock The pile: An 800gb dataset of diverse text for language modeling.
\newblock \emph{arXiv preprint arXiv:2101.00027}.

\bibitem[{Georganas et~al.(2025)Georganas, Kalamkar, Kozlov, and Heinecke}]{georganas2025ml}
Evangelos Georganas, Dhiraj Kalamkar, Alexander Kozlov, and Alexander Heinecke. 2025.
\newblock Ml-specqd: Multi-level speculative decoding with quantized drafts.
\newblock \emph{arXiv preprint arXiv:2503.13565}.

\bibitem[{Grattafiori et~al.(2024)Grattafiori, Dubey, Jauhri, Pandey, Kadian, Al-Dahle, Letman, Mathur, Schelten, Vaughan et~al.}]{grattafiori2024llama}
Aaron Grattafiori, Abhimanyu Dubey, Abhinav Jauhri, Abhinav Pandey, Abhishek Kadian, Ahmad Al-Dahle, Aiesha Letman, Akhil Mathur, Alan Schelten, Alex Vaughan, and 1 others. 2024.
\newblock The llama 3 herd of models.
\newblock \emph{arXiv preprint arXiv:2407.21783}.

\bibitem[{Guo et~al.(2025)Guo, Yang, Zhang, Song, Zhang, Xu, Zhu, Ma, Wang, Bi et~al.}]{guo2025deepseek}
Daya Guo, Dejian Yang, Haowei Zhang, Junxiao Song, Ruoyu Zhang, Runxin Xu, Qihao Zhu, Shirong Ma, Peiyi Wang, Xiao Bi, and 1 others. 2025.
\newblock Deepseek-r1: Incentivizing reasoning capability in llms via reinforcement learning.
\newblock \emph{arXiv preprint arXiv:2501.12948}.

\bibitem[{Hooper et~al.(2024)Hooper, Kim, Mohammadzadeh, Mahoney, Shao, Keutzer, and Gholami}]{hooper2024kvquant}
Coleman Hooper, Sehoon Kim, Hiva Mohammadzadeh, Michael~W Mahoney, Sophia Shao, Kurt Keutzer, and Amir Gholami. 2024.
\newblock Kvquant: Towards 10 million context length llm inference with kv cache quantization.
\newblock In \emph{Proceedings of NeurIPS}, pages 1270--1303.

\bibitem[{Kim et~al.(2024)Kim, Hooper, Gholami, Dong, Li, Shen, Mahoney, and Keutzer}]{kim2024squeezellm}
Sehoon Kim, Coleman Richard~Charles Hooper, Amir Gholami, Zhen Dong, Xiuyu Li, Sheng Shen, Michael~W Mahoney, and Kurt Keutzer. 2024.
\newblock Squeezellm: Dense-and-sparse quantization.
\newblock In \emph{Proceedings of ICML}, pages 23901--23923.

\bibitem[{Kwon et~al.(2023)Kwon, Li, Zhuang, Sheng, Zheng, Yu, Gonzalez, Zhang, and Stoica}]{kwon2023efficient}
Woosuk Kwon, Zhuohan Li, Siyuan Zhuang, Ying Sheng, Lianmin Zheng, Cody~Hao Yu, Joseph Gonzalez, Hao Zhang, and Ion Stoica. 2023.
\newblock Efficient memory management for large language model serving with pagedattention.
\newblock In \emph{Proceedings of SOSP}, pages 611--626.

\bibitem[{Leviathan et~al.(2023)Leviathan, Kalman, and Matias}]{leviathan2023fast}
Yaniv Leviathan, Matan Kalman, and Yossi Matias. 2023.
\newblock Fast inference from transformers via speculative decoding.
\newblock In \emph{Proceedings of ICML}, pages 19274--19286.

\bibitem[{Li et~al.(2024{\natexlab{a}})Li, Wei, Zhang, and Zhang}]{li2024eagle}
Yuhui Li, Fangyun Wei, Chao Zhang, and Hongyang Zhang. 2024{\natexlab{a}}.
\newblock Eagle-2: Faster inference of language models with dynamic draft trees.
\newblock In \emph{Proceedings of EMNLP}, pages 7421--7432.

\bibitem[{Li et~al.(2024{\natexlab{b}})Li, Wei, Zhang, and Zhang}]{li2024eagle1}
Yuhui Li, Fangyun Wei, Chao Zhang, and Hongyang Zhang. 2024{\natexlab{b}}.
\newblock Eagle: Speculative sampling requires rethinking feature uncertainty.
\newblock In \emph{Proceedings of ICML}, pages 28935--28948.

\bibitem[{Li et~al.(2025)Li, Wei, Zhang, and Zhang}]{li2025eagle}
Yuhui Li, Fangyun Wei, Chao Zhang, and Hongyang Zhang. 2025.
\newblock Eagle-3: Scaling up inference acceleration of large language models via training-time test.
\newblock \emph{arXiv preprint arXiv:2503.01840}.

\bibitem[{Lin et~al.(2024{\natexlab{a}})Lin, Tang, Tang, Yang, Chen, Wang, Xiao, Dang, Gan, and Han}]{lin2024awq}
Ji~Lin, Jiaming Tang, Haotian Tang, Shang Yang, Wei-Ming Chen, Wei-Chen Wang, Guangxuan Xiao, Xingyu Dang, Chuang Gan, and Song Han. 2024{\natexlab{a}}.
\newblock Awq: Activation-aware weight quantization for on-device llm compression and acceleration.
\newblock In \emph{Proceedings of MLSys}, pages 87--100.

\bibitem[{Lin et~al.(2024{\natexlab{b}})Lin, Tang, Yang, Zhang, Xiao, Gan, and Han}]{lin2024qserve}
Yujun Lin, Haotian Tang, Shang Yang, Zhekai Zhang, Guangxuan Xiao, Chuang Gan, and Song Han. 2024{\natexlab{b}}.
\newblock Qserve: W4a8kv4 quantization and system co-design for efficient llm serving.
\newblock \emph{arXiv preprint arXiv:2405.04532}.

\bibitem[{Liu et~al.(2024)Liu, Yuan, Jin, Zhong, Xu, Braverman, Chen, and Hu}]{liu2024kivi}
Zirui Liu, Jiayi Yuan, Hongye Jin, Shaochen Zhong, Zhaozhuo Xu, Vladimir Braverman, Beidi Chen, and Xia Hu. 2024.
\newblock Kivi: A tuning-free asymmetric 2bit quantization for kv cache.
\newblock In \emph{Proceedings of ICML}, pages 32332--32344.

\bibitem[{Merity et~al.(2016)Merity, Xiong, Bradbury, and Socher}]{merity2016pointer}
Stephen Merity, Caiming Xiong, James Bradbury, and Richard Socher. 2016.
\newblock Pointer sentinel mixture models.
\newblock In \emph{Proceedings of ICLR}.

\bibitem[{Patterson(2004)}]{patterson2004latency}
David~A Patterson. 2004.
\newblock Latency lags bandwith.
\newblock \emph{Communications of the ACM}, 47(10):71--75.

\bibitem[{Sadhukhan et~al.(2025)Sadhukhan, Chen, Chen, Tiwari, Lai, Shi, Yen, May, Chen, and Chen}]{sadhukhan2025magicdec}
Ranajoy Sadhukhan, Jian Chen, Zhuoming Chen, Vashisth Tiwari, Ruihang Lai, Jinyuan Shi, Ian En-Hsu Yen, Avner May, Tianqi Chen, and Beidi Chen. 2025.
\newblock Magicdec: Breaking the latency-throughput tradeoff for long context generation with speculative decoding.
\newblock In \emph{Proceedings of ICLR}.

\bibitem[{Shao et~al.(2024)Shao, Chen, Zhang, Xu, Zhao, Li, Zhang, Gao, Qiao, and Luo}]{shao2024omniquant}
Wenqi Shao, Mengzhao Chen, Zhaoyang Zhang, Peng Xu, Lirui Zhao, Zhiqian Li, Kaipeng Zhang, Peng Gao, Yu~Qiao, and Ping Luo. 2024.
\newblock Omniquant: Omnidirectionally calibrated quantization for large language models.
\newblock In \emph{Proceedings of ICLR}.

\bibitem[{Shazeer(2019)}]{shazeer2019fast}
Noam Shazeer. 2019.
\newblock Fast transformer decoding: One write-head is all you need.
\newblock \emph{arXiv preprint arXiv:1911.02150}.

\bibitem[{Sun et~al.(2024{\natexlab{a}})Sun, Chen, Yang, Tian, and Chen}]{sun2024triforce}
Hanshi Sun, Zhuoming Chen, Xinyu Yang, Yuandong Tian, and Beidi Chen. 2024{\natexlab{a}}.
\newblock Triforce: Lossless acceleration of long sequence generation with hierarchical speculative decoding.
\newblock In \emph{Proceedings of COLM}.

\bibitem[{Sun et~al.(2024{\natexlab{b}})Sun, Liu, Bai, Bao, Zhao, Li, Hu, Yu, Hou, Yuan et~al.}]{sun2024flatquant}
Yuxuan Sun, Ruikang Liu, Haoli Bai, Han Bao, Kang Zhao, Yuening Li, Jiaxin Hu, Xianzhi Yu, Lu~Hou, Chun Yuan, and 1 others. 2024{\natexlab{b}}.
\newblock Flatquant: Flatness matters for llm quantization.
\newblock \emph{arXiv preprint arXiv:2410.09426}.

\bibitem[{Tiwari et~al.(2025)Tiwari, Xi, Tomar, Hooper, Kim, Horton, Najibi, Mahoney, Keutzer, and Gholami}]{tiwari2025quantspec}
Rishabh Tiwari, Haocheng Xi, Aditya Tomar, Coleman Hooper, Sehoon Kim, Maxwell Horton, Mahyar Najibi, Michael~W Mahoney, Kurt Keutzer, and Amir Gholami. 2025.
\newblock Quantspec: Self-speculative decoding with hierarchical quantized kv cache.
\newblock \emph{arXiv preprint arXiv:2502.10424}.

\bibitem[{Tseng et~al.(2024)Tseng, Chee, Sun, Kuleshov, and De~Sa}]{tseng2024quip}
Albert Tseng, Jerry Chee, Qingyao Sun, Volodymyr Kuleshov, and Christopher De~Sa. 2024.
\newblock Quip $\# $: Even better llm quantization with hadamard incoherence and lattice codebooks.
\newblock In \emph{Proceedings of ICML}, pages 48630--48656.

\bibitem[{Xia et~al.(2024)Xia, Yang, Dong, Wang, Li, Ge, Liu, Li, and Sui}]{xia-etal-2024-unlocking}
Heming Xia, Zhe Yang, Qingxiu Dong, Peiyi Wang, Yongqi Li, Tao Ge, Tianyu Liu, Wenjie Li, and Zhifang Sui. 2024.
\newblock Unlocking efficiency in large language model inference: A comprehensive survey of speculative decoding.
\newblock In \emph{Findings of the ACL}, pages 7655--7671.

\bibitem[{Xiao et~al.(2023)Xiao, Lin, Seznec, Wu, Demouth, and Han}]{xiao2023smoothquant}
Guangxuan Xiao, Ji~Lin, Mickael Seznec, Hao Wu, Julien Demouth, and Song Han. 2023.
\newblock Smoothquant: Accurate and efficient post-training quantization for large language models.
\newblock In \emph{Proceedings of ICML}, pages 38087--38099.

\bibitem[{Zhang et~al.(2024)Zhang, Zhang, Huang, Xiang, Wang, Wang, Zhang, Yu, Liu, and Lin}]{zhang2024qqq}
Ying Zhang, Peng Zhang, Mincong Huang, Jingyang Xiang, Yujie Wang, Chao Wang, Yineng Zhang, Lei Yu, Chuan Liu, and Wei Lin. 2024.
\newblock Qqq: Quality quattuor-bit quantization for large language models.
\newblock \emph{arXiv preprint arXiv:2406.09904}.

\bibitem[{Zhao et~al.(2024)Zhao, Lu, Wang, Kong, and Wu}]{zhao2024qspec}
Juntao Zhao, Wenhao Lu, Sheng Wang, Lingpeng Kong, and Chuan Wu. 2024.
\newblock Qspec: Speculative decoding with complementary quantization schemes.
\newblock \emph{arXiv preprint arXiv:2410.11305}.

\bibitem[{Zhao et~al.(2025)Zhao, Pan, Han, Zhang, Sun, Huang, Zhang, Zhao, Li, Wang et~al.}]{zhao2025fr}
Weilin Zhao, Tengyu Pan, Xu~Han, Yudi Zhang, Ao~Sun, Yuxiang Huang, Kaihuo Zhang, Weilun Zhao, Yuxuan Li, Jianyong Wang, and 1 others. 2025.
\newblock Fr-spec: Accelerating large-vocabulary language models via frequency-ranked speculative sampling.
\newblock \emph{arXiv preprint arXiv:2502.14856}.

\bibitem[{Zheng et~al.(2023)Zheng, Chiang, Sheng, Zhuang, Wu, Zhuang, Lin, Li, Li, Xing et~al.}]{zheng2023judging}
Lianmin Zheng, Wei-Lin Chiang, Ying Sheng, Siyuan Zhuang, Zhanghao Wu, Yonghao Zhuang, Zi~Lin, Zhuohan Li, Dacheng Li, Eric Xing, and 1 others. 2023.
\newblock Judging llm-as-a-judge with mt-bench and chatbot arena.
\newblock In \emph{Proceedings of NeurIPS}, pages 46595--46623.

\end{thebibliography}

\newpage
\appendix

\section{EAGLE-3 Performance}
\label{sec:appendix}

\begin{table*}[t]
  \centering
  \setlength{\tabcolsep}{4pt}
  \scalebox{0.75}{
    \begin{tabular}{lc|cccccccccccc|cc}
      \toprule
      \multirow{2}{*}{\textbf{Method}} 
    & \multirow{2}{*}{$d$}
      & \multicolumn{2}{c}{\textbf{MT.}}
      & \multicolumn{2}{c}{\textbf{Conv.}}
      & \multicolumn{2}{c}{\textbf{RAG}}
      & \multicolumn{2}{c}{\textbf{Math}}
      & \multicolumn{2}{c}{\textbf{QA}}
      & \multicolumn{2}{c}{\textbf{Summ.}}
      & \multicolumn{2}{|c}{\textbf{Avg.}} \\
      \cmidrule(lr){3-4}\cmidrule(lr){5-6}\cmidrule(lr){7-8}\cmidrule(lr){9-10}\cmidrule(lr){11-12}\cmidrule(lr){13-14}\cmidrule(l){15-16}
& 
& $\tau$ & \textbf{Tok/s}
& $\tau$ & \textbf{Tok/s}
& $\tau$ & \textbf{Tok/s}
& $\tau$ & \textbf{Tok/s}
& $\tau$ & \textbf{Tok/s}
& $\tau$ & \textbf{Tok/s}
& $\tau$ & \textbf{Tok/s} \\
\midrule
Vanilla AR & -
& 1.00 & 36.11
& 1.00 & 34.90
& 1.00 & 34.82
& 1.00 & 35.41
& 1.00 & 35.57
& 1.00 & 34.41
& 1.00 & 35.20 (1.0 $\times$) \\
\midrule
\multirow{2}{*}{Vanilla SP}
& 6
& 4.23 & 77.31
& 4.60 & 76.40
& 4.44 & 77.05
& 5.39 & 91.87
& 4.23 & 72.84
& 4.26 & 68.30
& 4.54 & 77.30 (2.20$\times$) \\
& 7
& 4.48 & 75.11
& 4.95 & 75.82
& 4.72 & 75.98
& 5.90 & 92.76
& 4.49 & 71.54
& 4.55 & 67.14
& 4.87 & 76.39 (2.17$\times$) \\

\midrule
\multirow{4}{*}{EAGLE-3} 
& 3 
& 3.30 & 87.63
& 3.54 & 88.72
& 3.47 & 87.42
& 3.69 & 94.98
& 3.31 & 86.00
& 3.35 & 83.75
& 3.48 & 88.08 (2.50$\times$) \\
& 4
& 3.74 & 95.90
& 4.12 & 98.86
& 4.04 & 98.55
& 4.31 & 108.02
& 3.75 & 93.55
& 3.83 & 91.05
& 4.02 & 97.65 (2.77$\times$) \\
& 5
& 4.05 & 85.30
& 4.60 & 90.78
& 4.48 & 90.73
& 4.95 & 101.47
& 4.13 & 84.96
& 4.17 & 82.47
& 4.47 & 89.28 (2.54$\times$) \\
& 6
& 4.20 & 85.60
& 4.95 & 95.49
& 4.81 & 95.09
& 5.39 & 107.50
& 4.39 & 87.41
& 4.44 & 85.09
& 4.79 & 92.70 (2.63$\times$) \\
\midrule

\multirow{4}{*}{\textbf{HierSpec}}
& 6(3)
& 4.60 & 94.16
& 5.23 & 101.86
& 4.92 & 99.65
& 6.41 & 126.64
& 4.65 & 91.29
& 4.71 & 87.12
& 5.11 & \textbf{100.12 (2.84$\times$)} \\
& 6(4)
& 4.59 & 90.22
& 5.03 & 94.91
& 4.83 & 94.30
& 6.01 & 118.47
& 4.61 & 86.21
& 4.68 & 82.45
& 4.97 & 94.43 (2.68$\times$) \\
& 7(3)
& 4.66 & 92.08
& 5.30 & 100.92
& 5.04 & 99.60
& 6.55 & 127.05
& 4.76 & 90.43
& 4.83 & 85.97
& 5.21 & 99.34 (2.82$\times$) \\
& 7(4)
& 4.81 & 88.50
& 5.59 & 100.28
& 5.25 & 100.49
& 7.00 & 127.76
& 4.90 & 86.57
& 4.98 & 82.58
& 5.45 & 97.69 (2.78$\times$) \\

\bottomrule       
    \end{tabular}
  }
  \caption{Average accepted length $\tau$ and decoding speed \textbf{Tok/s} of different methods on W4A16 Llama-3.3-70B with different draft lengths $d$. The numbers in draft length parentheses (e.g., 3 in 6(3)) denote the draft length of EAGLE-3 for the small model W4A16 Llama-3.1-8B in the first hierarchy. And the numbers in decoding speed parentheses ($2.84\times$) represent the speedup compared to the W4A16 vanilla auto-regressive approach.}
  \label{tab:hspec_experiment_ea3}
\end{table*}

Although the lack of public training details hinders the integration of EAGLE-3 into our hierarchical framework, we assess HierSpec with publicly available EAGLE-3 checkpoints.
For EAGLE-3, we adopt Llama-3.3-70B-Instruct as the target model, which is the only 70B size model with an available EAGLE-3 checkpoint.
For HierSpec, due to the absence of the 8B size model in Llama-3.3 series, we select W4A16 Llama-3.1-8B-Instruct as the small intermediate model.
Even under such an unfair configuration, HierSpec achieves a further speedup over our highly optimized EAGLE-3 implementation as presented in Table~\ref{tab:hspec_experiment_ea3}.
Under the same draft length $d=6$, the average accepted length of Vanilla SP is smaller than EAGLE-3.
And the average accepted length of HierSpec shows an obvious degradation on some tasks compared to the HierSpec for W4A16 Llama-3-70B-Instruct.
This suggests that the inferior alignment between the intermediate small model Llama-3.1-8B-Instruct and the target model Llama-3.3-70B-Instruct limits the potential of our HierSpec framework.

\end{document}